%% file: main.tex
\documentclass[11pt]{article}

\usepackage[utf8]{inputenc}
\usepackage[T1]{fontenc}

\usepackage{arxiv}

\usepackage{amsmath,amssymb,amsfonts}
\usepackage{graphicx}
\usepackage{float}
\usepackage{afterpage}
\usepackage{booktabs}
\usepackage{multirow}
\usepackage{natbib}
\usepackage{enumitem}
\usepackage{subcaption}
\usepackage{array}
\usepackage{longtable}
\usepackage{tabularx}
\usepackage{lscape}
\usepackage{pdflscape}

\usepackage{hyperref}
\usepackage{xurl}
\newcommand{\repodoc}[2]{\href{https://github.com/19PINE-AI/namerank/blob/main/docs/#1}{#2}}

\hypersetup{colorlinks=true, linkcolor=arxivaccent, citecolor=arxivaccent, urlcolor=arxivaccent, breaklinks=true}
\runningtitle{The Model Knows Your Project, Not You: Measuring Recognition in LLMs with NameRank}
\title{The Model Knows Your Project, Not You: Measuring Recognition in LLMs with NameRank}
\author{%
  \begin{tabular}{@{}c@{\hspace{4em}}c@{}}
    Bojie Li & Noah Shi \\
    Pine AI & University of Washington
  \end{tabular}%
}
\date{}

\begin{document}
\maketitle

\begin{abstract}
What a frontier model recalls about a person or tool from its own weights---before any retrieval step---often shapes the first description a human sees, making that parametric corpus presence a measurement problem. Citations explain about a third of whether a model recognizes a researcher; we target the residual and build \textbf{NameRank}, a $[0,1]$ recognition score: each of $4{,}685$ entities in $54$ cohorts is probed with one open-ended question across $36$ models, and an independent judge returns a binary verdict against a curated gold---did the model state a specific, non-guessable fact about this exact entity?---so hallucination, context echo, and guesses earn nothing. Synthetic-null entities hold the floor near zero, and verdicts track the entity, not the model. One thesis organizes the findings: \emph{recognition is paid to named, indexable artifacts, not to credentials or titles.} Every Olympic-style credential sits below a working-researcher baseline, because no named artifact ships with the medal, yet the ranking inverts at the marquee tier, where Nobel, Turing, and Fields laureates saturate the panel. For independent creators the tool out-ranks its maker, and the credential that \emph{does} propagate is a named method or awarded paper. Being one of many named contributors to a celebrated artifact, by contrast, earns almost nothing---the authors listed on a flagship model report or system card sit near the recognition floor---because recognition attaches to the artifact's own distinctive name, not to the roster behind it. No bibliometric predicts recognition well; top-density institutions out-recognize peers at matched citations; and on $258$ news events recognition loads on peak salience, not persistence. A self-report probe shows introspection reads a corpus prior, not its own knowledge.
\end{abstract}
\begin{center}
\small
Code: \url{https://github.com/19PINE-AI/namerank} \\[2pt]
Website: \url{https://01.me/research/namerank}
\end{center}
\vspace{-0.6em}

\section{Introduction}
\label{sec:intro}

\begin{figure}[p]
    \centering
    \includegraphics[width=0.86\textwidth]{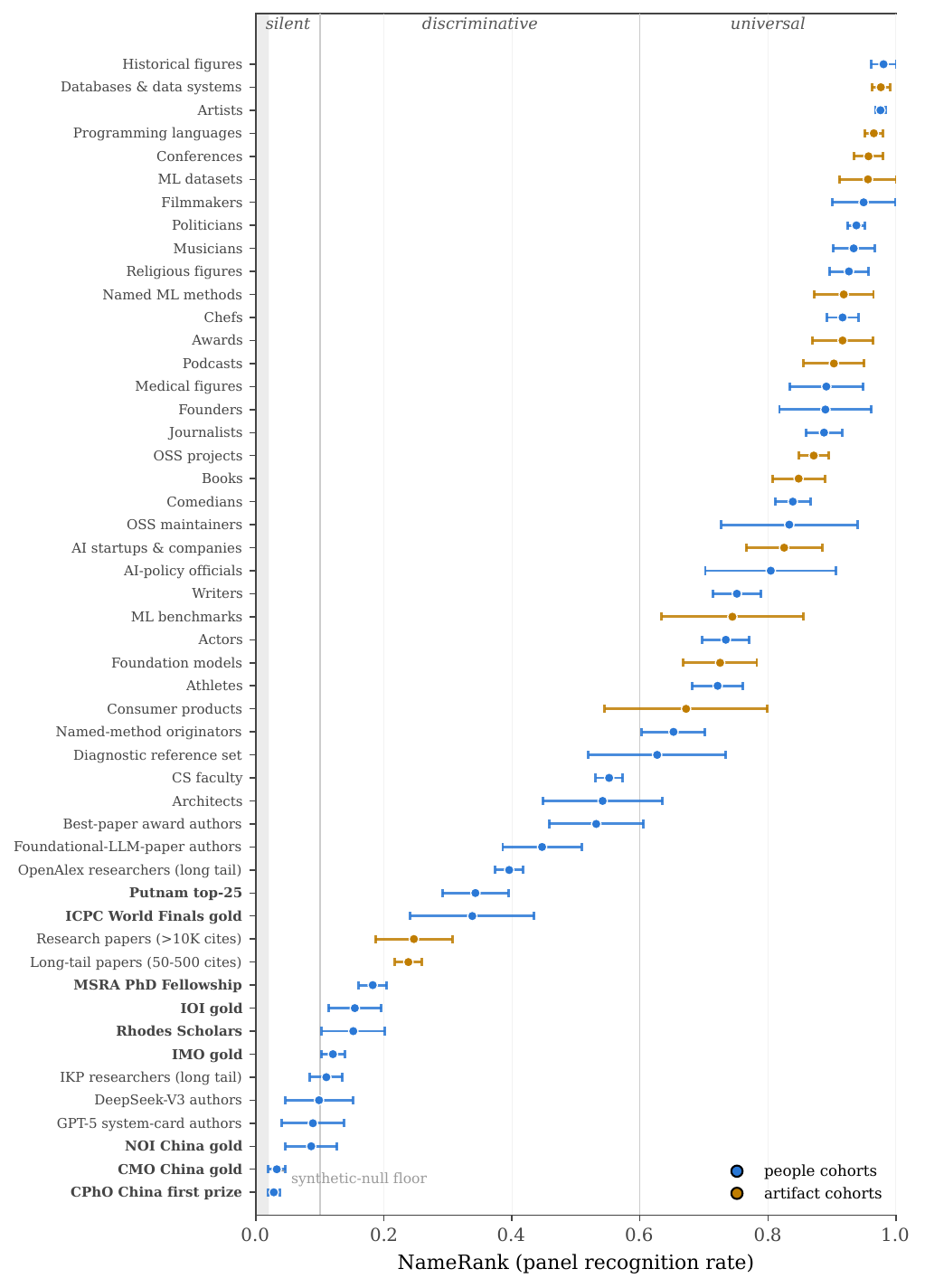}
    \caption{\textbf{Every cohort placed on one recognition scale.} Each cohort ($n \geq 10$) on the NameRank axis: dot is the mean panel recognition rate, whisker the $95\%$ CI; people cohorts in blue, artifacts in amber. The hatched band is the synthetic-null floor; olympiad and fellowship credentials in bold. Three zones are labelled: \emph{silent}, \emph{discriminative}, \emph{universal} (Section~\ref{sec:results-headline}).}
    \label{fig:overview}
\end{figure}

Ask a frontier language model about an International Mathematical Olympiad gold medalist and, drawing only on what it has internalized, it will usually answer ``unknown''---only about one model in eight recognizes them---while a popular open-source tool is described in detail by nearly every model on a $36$-model panel. A credential that selects a few hundred people a year is committed to memory less reliably than a well-named side project. This matters to the extent that human curiosity about people and their work is now routed first through a model---a premise, not a result: a practitioner deciding whether to read a paper, hire a candidate, or invest in a startup increasingly begins with what a frontier model says. Such systems can reach for web search, so parametric memory is not the whole of what a user sees; but it sets the default---what the model volunteers before any tool fires, and what it returns when retrieval is unavailable or untriggered---where a name's presence in the weights separates a coherent summary from an ``unknown'' or a confabulated biography. As this first pass shifts from search rankings, conference plenaries, and citation counts toward a small number of frontier models, whether a name has reached their weights becomes a substantive measurement problem---the one we take up here, holding retrieval-augmented behavior outside our scope.

Bibliometrics do not solve this measurement problem. In earlier work on the Incompressible Knowledge Probes (IKP) benchmark, \citet{li2026ikp} found---as a byproduct of calibrating the knowledge capacity of frontier models---that citation counts explain only about a third of the variance in whether a model recognizes a researcher. A controlled audit attributed the remainder to three corpus-level mechanisms: \emph{name uniqueness}, \emph{named-artifact attachment}, and \emph{subfield-ecosystem density}. That work characterized the residual but did not measure it: its recognition scores live on a coarse tier ladder built for parameter estimation, not for comparing entities. We take the opposite stance---the residual is the quantity of interest---and build a dedicated instrument for it. Section~\ref{sec:related} details how the instrument departs from its predecessor.

NameRank measures \emph{per-query presence}: conditional on a human asking a model about an entity, whether the model actually recognizes it. It is deliberately not weighted by how \emph{often} an entity is queried---that human-facing exposure dimension is real but outside our scope and we do not measure it---so the score isolates the cross-entity variation in corpus presence rather than reach. Nor do we claim it measures \emph{quality} or \emph{merit}; it measures whether the corpora that mediate human curiosity at scale have internalized the entity's name.

\textbf{NameRank} is a $[0,1]$ recognition score. Each entity is probed with one fixed open-ended question (``tell me what you know about [name], who [context]'') across a panel of $36$ frontier models. An independent LLM judge reads every response against a curated gold answer and returns a binary recognition verdict---true only when the response states a specific, non-guessable fact that is correct and \emph{not} derivable from the disambiguating context, so a fluent hallucination, a restatement of the context, or a lucky guess contributes nothing. The per-entity score is the panel recognition rate: the probability that a randomly drawn frontier model genuinely knows the entity. Three properties distinguish this from recognition numbers that fall out of benchmark suites: it places researchers, founders, open-source maintainers, and named artifacts on a single axis; it is hallucination-resistant by construction; and cross-model disagreement is retained as signal rather than averaged away.

A single thesis runs through the results and organizes the paper. \emph{Recognition in the LLM channel is paid to named, indexable artifacts---not to credentials, titles, or reach.} A credential propagates a name only to the extent that named production attaches to it, and one axis makes this plain: an olympiad medal, which ships with no named artifact, sits near the floor; a best-paper award, which ships with a named paper, sits well above the working-researcher baseline; authorship of a widely-used named method sits higher still; and the marquee prizes, backed by a career of named production, sit at the top against the panel ceiling. The artifact-over-creator inversion---where a tool out-ranks its maker and no prompt can manufacture the reverse---is the mechanism seen most directly; the institution gradient and the news-event calibration are the same mechanism from other angles.

\paragraph{Contributions.}
\begin{enumerate}[leftmargin=*,itemsep=1pt]
    \item A \textbf{cross-domain recognition instrument}---a binary recognition verdict over a $36$-model panel, calibrated by synthetic-null floors---that places researchers, founders, OSS authors, public websites, and named artifacts on one $[0,1]$ axis (Section~\ref{sec:method}, \ref{sec:results-headline}).
    \item The \textbf{career-arc account of credentials}: every Olympic-style credential scores at or below a working-researcher baseline, while best-paper awards, named-method authorship, and marquee prizes rise above it---a unified picture in which a credential propagates a name in proportion to the named artifacts it carries (Sections~\ref{sec:results-credentials}--\ref{sec:results-awards}).
    \item The \textbf{named-artifact mechanism}: for independent creators the artifact out-ranks its maker; the effect is an observational property of corpus attribution, not something a context intervention can manufacture, and it rules out the intuitive reading of recognition as reach (Section~\ref{sec:results-mechanism}).
    \item \textbf{What predicts recognition}: no bibliometric stands in for recognition---$h$-index, citations, and attribution-weighted variants all explain only about a fifth of the variance; a corpus-density gradient runs across institutions (even at matched citations) and countries; and a $258$-event news cohort with a recorded attention ledger calibrates the instrument externally---peak salience, not persistence (Sections~\ref{sec:results-external}--\ref{sec:results-events}).
    \item A \textbf{validation battery} over wording, vintage, judge family, synthetic nulls, and gold confounds, plus a \textbf{self-report probe} showing introspection reads the corpus prior rather than the model's own state (Sections~\ref{sec:results-validation}--\ref{sec:results-selfreport}); and an \textbf{open release} of all probes, golds, responses, and code.
\end{enumerate}

\section{Measuring NameRank}
\label{sec:method}

\begin{figure}[!htbp]
    \centering
    \includegraphics[width=\textwidth]{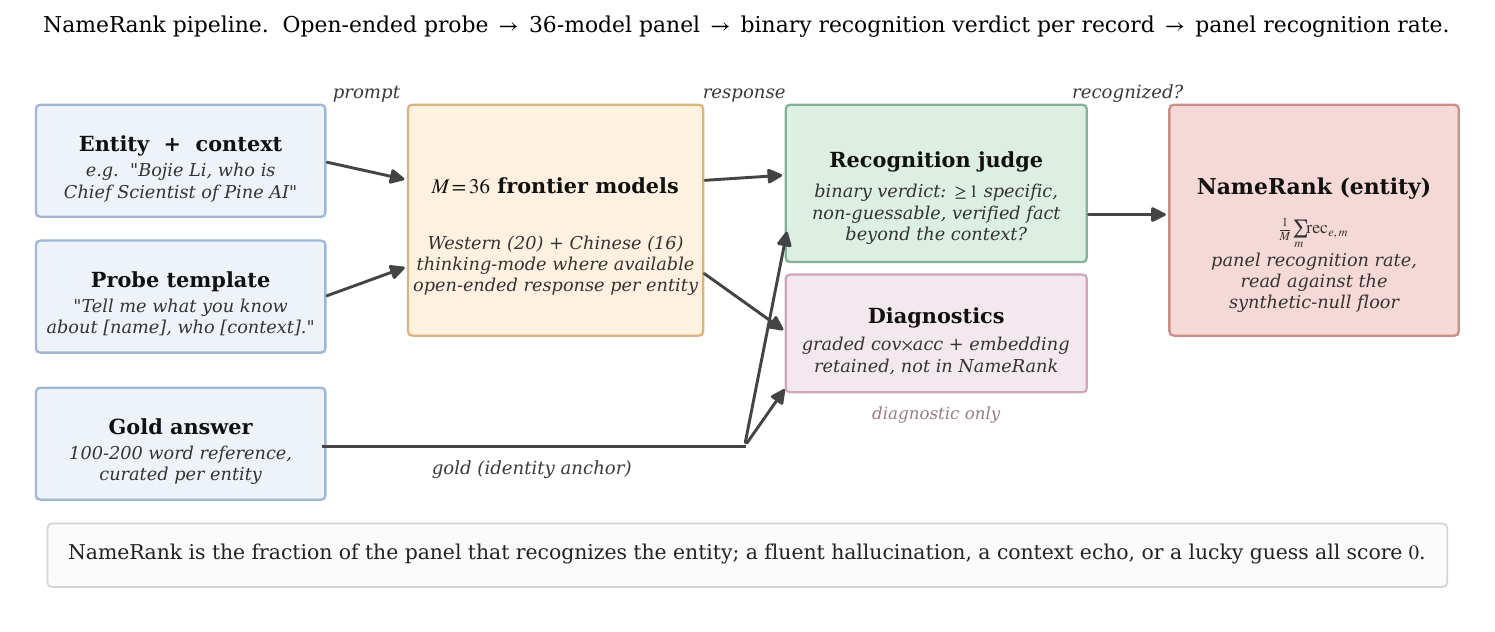}
    \caption{NameRank pipeline. Each of the $36$ panel models answers a fixed open-ended probe (entity name plus a short role/field context); an LLM judge scores the response against a gold answer, crediting recognition only for a specific, non-guessable, correct fact not derivable from the context. The per-entity score is the panel recognition rate. Embedding similarity and a graded coverage$\times$accuracy score are stored as diagnostics only (Appendix~\ref{app:scoring}).}
    \label{fig:pipeline}
\end{figure}

Figure~\ref{fig:pipeline} summarizes the pipeline this section unpacks: a fixed open-ended probe is sent to a $36$-model panel for each entity; an independent LLM judge reads each (entity, model) response against a curated gold answer and returns a binary \emph{recognition} verdict---did the response demonstrate genuine memory of this specific entity; and the per-entity NameRank is the fraction of the panel that recognizes it.

\subsection{Definition}
\label{sec:construct}

For an entity $e$ with a disambiguating context $c(e)$ and a curated gold answer $g(e)$, and a model panel $\mathcal{M}$ of size $M = 36$, NameRank is the fraction of the panel that recognizes the entity,
\begin{equation}
\mathrm{NameRank}(e) \;=\; \frac{1}{M} \sum_{m \in \mathcal{M}} \mathrm{rec}(r_{e,m},\, c(e),\, g(e)),
\label{eq:namerank}
\end{equation}
where $r_{e,m}$ is the response of model $m$ to the open-ended probe
\begin{quote}
\small ``\emph{Tell me what you know about [name], who [context]. If you do not recognize this entity, respond with `unknown'. Limit your response to about 150 words.}''
\end{quote}
and $\mathrm{rec} \in \{0,1\}$ is a recognition verdict produced by an LLM judge (Section~\ref{sec:method-judge}): it is $1$ only when the response states at least one \emph{specific, non-guessable} fact that is true of this exact entity---verified against the gold answer or the judge's own knowledge---and not derivable from the disambiguating context. A refusal, a hedged non-answer, a response confined to context-derivable generalities, and a fluent but wrong-person biography all score $0$. NameRank thus reads as the probability that a randomly drawn frontier model genuinely knows the entity---concretely, a NameRank of $0.12$ means about four of the $36$ models recognized the entity while the rest refused or hallucinated, and $0.90$ means essentially all did. Three worked (entity, model, response) triples spanning the range are provided in the \repodoc{worked-examples.md}{repository documentation}.

\subsection{Probe Design}
\label{sec:probe}

\textbf{One open-ended probe per entity.} Closed-factoid probes have a known failure mode: a model that recognizes the entity through fact $X$ but does not recall fact $Y$ scores zero on a probe-of-$Y$, biasing the metric toward specific-fact recall. The open-ended probe instead asks the holistic question: does the model have a retrievable representation of this entity? The template is fixed, with two slots: the canonical entity name, and a short disambiguating clause giving role, affiliation, and field (e.g., \emph{``Wei Zhang, who is a professor of mathematics at MIT working on PDEs''}). The context resolves name collisions at probe time without revealing the gold answer. The full English and Chinese probe text and the judge prompt are reproduced verbatim in the \repodoc{probe-templates.md}{repository documentation}.

The context names role, affiliation, and field---never specific contributions, papers, dates, or named artifacts---so it anchors disambiguation without leaking the gold; the judge's rubric credits only specific named facts, and generic in-context biographies earn nothing (evidence in Appendix~\ref{app:spec}; a dedicated context-ablation in Appendix~\ref{app:confounds}).

\subsection{Gold Answers}
\label{sec:gold}

For each entity we construct a gold answer built from \emph{entity-specific, verifiable facts}---named works, named contributions, real affiliations with dates, and other identifying particulars---composed from primary and authoritative sources: bibliographic records (OpenAlex, Semantic Scholar, Crossref) for researchers and papers, official competition and award records for credential holders, code-host metadata and READMEs for open-source artifacts, encyclopedia lead sections for public figures and companies, and length-normalized web-grounded profiles for the long tail. The gold serves two roles for the judge: it \emph{identifies} which specific entity the probe refers to (disambiguating same-name individuals), and it supplies a correctness anchor. It deliberately does \emph{not} restate the disambiguating context: a fact the probe already supplies earns no recognition credit, so the gold's discriminating content is the material a model must know independently. We audit a stratified sample against primary sources following the protocol of \citet{li2026ikp}; the observed correction rate is small, typically year-of-affiliation drift. A confound check confirms NameRank does not reduce to a has-a-Wikipedia-page flag (Appendix~\ref{app:confounds}).

\subsection{The Judge}
\label{sec:method-judge}

We use Gemini~3 Flash Preview as the judge (direct Google API to minimize routing variance; temperature $0$). The judge is shown the entity name, the probe context, the gold answer, and the model's response, and returns a binary \emph{recognized} verdict. It is instructed to credit recognition only when the response states a specific, non-guessable fact that is (a)~true of this exact entity, established either by the gold answer or by the judge's own reliable knowledge; (b)~not derivable from the probe context; and (c)~about the same entity the gold identifies, not a same-name individual. The judge uses knowledge \emph{asymmetrically}: it may refute freely---flagging a wrong-person biography or a false claim using anything it knows---but it may \emph{credit} a fact only when it positively verifies it, so a plausible-sounding but unverified specific about an entity it does not recognize counts for nothing (the signature of hallucination on obscure or fictional names). For competition participants, fellowship recipients, and minor-paper authors---where an LLM judge has no reliable per-individual memory---an anti-confabulation provision restricts credit to facts present in the gold or to major, widely-documented achievements, never to fine-grained particulars asserted from the judge's own memory. Broad field, nationality, an employer inferable from the name, generic praise, and restatements of the context are all non-credit by construction.

We also record, per response, a graded coverage$\times$accuracy score and the embedding cosine similarity to the gold~\citep{bge2024}; both are retained as diagnostics and neither enters the headline NameRank. Appendix~\ref{app:scoring} shows why a judge is required---embedding similarity credits fluent hallucinations as recognized---and why the recognition verdict is preferred to a graded score: graded coverage measures overlap between two arbitrary samples of an entity's fact universe, penalizing a model that knows the entity through facts the gold happens to omit, whereas recognition asks the construct-faithful question of whether the model knows the entity at all. The dependence of the findings on the judge's model family is quantified with a multi-judge re-grade in Appendix~\ref{app:cross-judge}.

\subsection{Aggregation}
\label{sec:aggregation}

NameRank is the unweighted panel recognition rate across the $36$-model panel (Equation~\ref{eq:namerank}). We also retain, per entity, the refusal rate, the Western/Chinese sub-panel rates, the diagnostic graded score, and the raw response text. Averaging across the panel is what makes the score stable: a single model contributes noise from prompt sensitivity, refusal-policy idiosyncrasies, and training-data accidents, while the panel rate integrates corpus presence over the current frontier fleet. The retained per-model verdicts preserve the residual that the panel mean averages over---per-model generosity and vendor-family differences, tabulated in the \repodoc{per-model-stats.md}{released per-model statistics}.

\begin{figure}[!htbp]
    \centering
    \includegraphics[width=\textwidth]{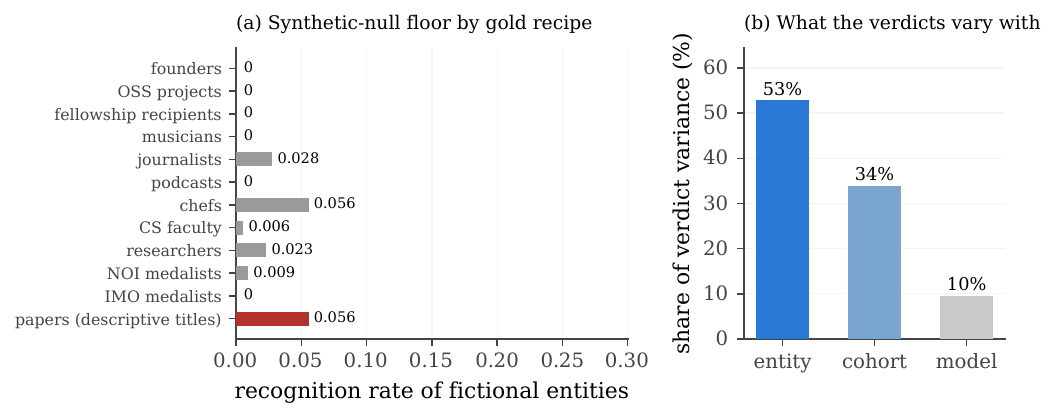}
    \caption{\textbf{The recognition metric is calibrated.} \textbf{(a)}~Fictional entities of each gold recipe earn near-zero recognition (except descriptive paper titles). \textbf{(b)}~Per-record verdicts vary with the entity, not the model.}
    \label{fig:calibration}
\end{figure}

\textbf{Calibration.} Two properties make the resulting axis trustworthy (Figure~\ref{fig:calibration}). First, a \emph{synthetic-null floor}: for each cohort we construct fictional entities on the same gold recipe---plausible but non-existent researchers, founders, medalists---and run them through the full pipeline. Because the judge credits only positively-verified facts, these earn essentially zero (${\leq}0.02$ for people and artifacts), so a nonzero score is real recognition rather than a guessing artifact; the one exception is fictional papers, whose descriptive titles admit plausible guesses ($0.06$), and this floor is subtracted where relevant. Second, a \emph{variance decomposition} of the per-record verdicts: the entity factor explains $56\%$ of the variance against $9\%$ for the model, so the score measures the entity, not the panel. Scores are read against the floor throughout (Appendix~\ref{app:synthetic}, \ref{app:scoring}).

\subsection{Panel and Cohorts}
\label{sec:panel}

The panel comprises $36$ frontier and near-frontier models accessed via OpenRouter and direct vendor APIs, spanning Western and Chinese vendors and frontier-reasoning to small (sub-$30$B) classes, with thinking mode enabled wherever a variant exists---the operational mode in which most user queries are answered (full roster in Appendix~\ref{app:spec}, Table~\ref{tab:panel}). The headline NameRank is the unweighted mean over all $36$ models; the Western/Chinese partition enters only in the cross-language robustness check (Section~\ref{sec:results-language}), where both model classes are shown to shift together on Chinese prompts.

We evaluate $4{,}685$ entities in $54$ core cohorts---credentialed individuals (olympiad medalists, Putnam, ICPC, Rhodes, MSRA fellows), working academics, industry figures, named artifacts, and mid-tier cross-profession figures (writers, athletes, actors, chefs, podcasts)---plus a hand-curated diagnostic reference set and special cohorts (technical-report and system-card author lists, conferences, awards); award, event, LLM-area, and university extensions are probed under the same protocol (per-group breakdown in Appendix~\ref{app:spec}, Table~\ref{tab:cohort-groups}). The design goal is coverage of the axes the metric claims to unify: credentialed individuals, whose recognition should be carried by the credential \emph{if} credentials propagate; working academics with known bibliometrics, as the external-validity anchor; industry figures whom no bibliometric measures; named artifacts, to compare things against their makers; and mid-tier figures, to test the axis outside the technology world. A $240$-entity subset was re-probed with a Chinese-translated template (Section~\ref{sec:results-language}).

\section{Recognition Across All Cohorts}
\label{sec:results}
\label{sec:results-headline}

\begin{figure}[!t]
    \centering
    \includegraphics[width=\textwidth]{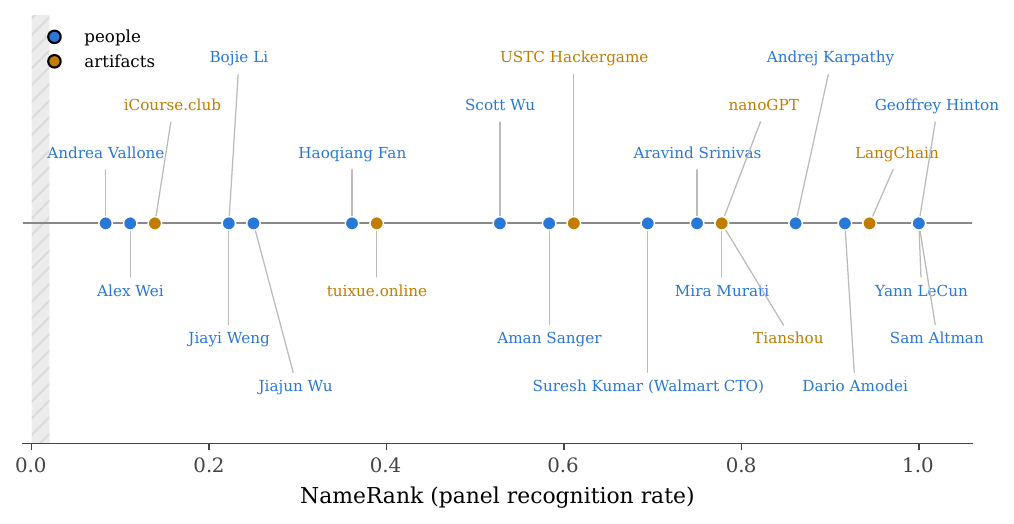}
    \caption{\textbf{Named entities from many domains on one recognition scale.} Selected named entities; people in blue, artifacts in amber. The hatched band is the synthetic-null floor.}
    \label{fig:axis-strip}
\end{figure}

Figure~\ref{fig:overview} places every cohort on the single recognition axis, and three properties of the instrument are visible at once. First, the axis is \emph{meaningful across domains}: researchers, founders, open-source tools, competition medalists, and named methods all land on one $[0,1]$ scale, something no bibliometric or domain-specific metric supports. Second, the score is \emph{silent, not misleading, below threshold}: the lowest cohorts sit inside the hatched synthetic-null band, where the panel withholds recognition rather than fabricating rankings. Third, an \emph{artifact/person split} is already visible---amber artifact cohorts cluster at the universal end while the bold olympiad credentials sit at the bottom.

Three zones organize the axis. The \emph{silent} zone (at the floor) holds names that have not propagated: GPT-5 system-card authors ($0.09$) and DeepSeek-V3 authors ($0.10$), each a list of real, accomplished engineers whose names appear only inside a large contributor roster. The \emph{discriminative} zone is where the instrument does most of its work: working researchers ($0.40$), CS faculty ($0.55$), and the credential cohorts spread across it. The \emph{universal} zone holds named artifacts and public figures---named ML methods ($0.92$), OSS projects ($0.87$), programming languages and AI companies near the ceiling---that essentially every model knows.

Figure~\ref{fig:axis-strip} makes the claim concrete. A Fortune-10 CTO, open-source tools, and community-run systems share one axis with Yann LeCun and Geoffrey Hinton (both $1.00$): citations miss the CTO, $h$-index misses nanoGPT's creator, GitHub stars miss Sam Altman---no legacy metric spans them, but recognition places them together.

\section{Credentials Predict Little Recognition}
\label{sec:results-credentials}

Every olympiad and fellowship credential sits at or below the working-researcher baseline. That pattern---and its inversion at the marquee tier---is the most visible consequence of a single mechanism: recognition attaches to named, indexable artifacts, not to the credential itself, so a credential propagates a name only to the extent that named production follows it.

\subsection{Nine elite credentials score below working researchers}

\begin{figure}[!htbp]
    \centering
    \includegraphics[width=0.82\textwidth]{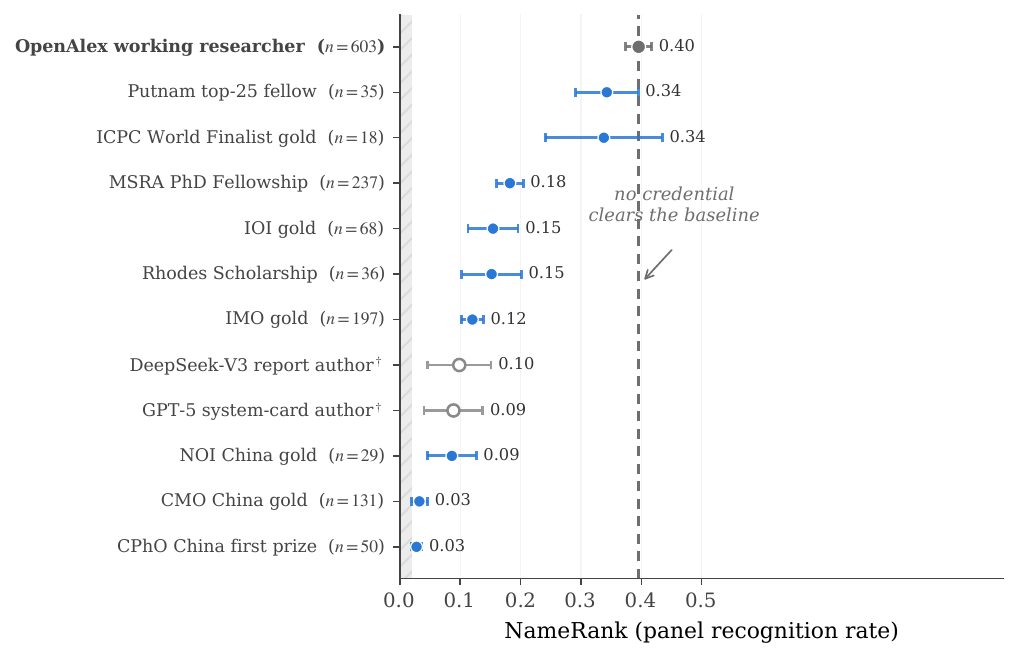}
    \caption{\textbf{Recognition of nine olympiad and fellowship credentials.} Each credential on the recognition axis (blue, $95\%$ CI), with the working-researcher baseline (dashed, $0.40$) and the synthetic-null floor (hatched, ${\sim}0.02$). $^\dagger$~Two large-team author-list cohorts (not credentials) shown for context.}
    \label{fig:credential-ladder}
\end{figure}

We place nine competition and fellowship credentials---each once a career-defining distinction---on the recognition axis against a baseline of ordinary citation-tracked researchers (OpenAlex, recognition $0.40$; Figure~\ref{fig:credential-ladder}). \emph{Every one falls below it.} The two nearest---Putnam Fellows and ICPC World Finalists ($0.34$ each)---are exactly the credentials that feed almost exclusively into US-tracked, English-corpus careers, and even they fall short; the globally-dispersing olympiads sit near the synthetic-null floor (IMO $0.12$, CMO and CPhO at $0.03$). Below even the credentials sit two large-team author-list cohorts (DeepSeek-V3 report, GPT-5 system card): membership in a famous project, absent an individually named artifact, propagates almost nothing.

Two properties confirm that the career, not the medal, carries the signal. First, \emph{career track dominates}: within IMO gold medalists, US-tracked academic and industry careers---the subset comparable to the citation-floored baseline---score materially higher than non-Anglophone-tracked golds, which is why Putnam and ICPC come closest to the baseline while IMO and CMO fall furthest below it. Second, \emph{recency does not matter}: medal year is essentially uncorrelated with recognition, so what accrues is post-medal production rather than the medal's freshness (Section~\ref{sec:discussion-clock}). The medal is not an exception to recognition decay but a proxy for selection into careers that later produce English text---and only the production propagates the name.

\textbf{The generational claim.} Olympic-style credential systems were optimized for a pre-LLM signaling regime in which a distinction translated into reputation through human social networks \citep{spence1973}. In the LLM-mediated regime, only credentials that translate into \emph{named, indexable production}---papers, tools, companies, talks carrying the holder's name---propagate; the medal itself, absent that production, does not. This dissolves a natural objection: that many prominent AI founders were olympiad medalists (the founders of Cognition, Perplexity, and Hyperliquid among them). They are recognized for what they \emph{built}, not for the medal, and the named artifact still out-ranks the person---Perplexity ($0.89$) sits above its founder Aravind Srinivas ($0.75$, no olympiad background), who in turn sits far above co-founder Johnny Ho ($0.28$), a three-time IOI gold medalist whose individual name reached the corpus far less than the company he helped build. Such founders are a selected few whose \emph{post-medal} production carried their names; across the full rosters the medal marks talent but propagates almost nothing on its own.

\subsection{Within one contest, medal rank does not predict recognition}
\label{sec:results-tianshou}

Probing a single contest's full roster isolates the credential from the career. On the complete China National Olympiad in Informatics 2009 roster---every gold, silver, and bronze medalist---the gold/non-gold cliff survives (gold recognition $0.10$ against $0.03$/$0.02$ for silver/bronze, Kruskal--Wallis $p<10^{-5}$, against a near-zero synthetic-null floor), while contest \emph{score} within a tier does not predict recognition at all. The few medalists who rise above the near-floor band are exactly those whose \emph{post-medal} careers produced named work---a now-prominent vision professor, a research lead, the author of this paper on a systems-and-startup track (Appendix~\ref{app:medal-tiers}). The medal grade sets the odds of entering a documented cohort; the recognition tracks what the person later built.

\subsection{Credentials that carry named work do raise recognition}
\label{sec:results-awards}

\begin{figure}[!htbp]
    \centering
    \includegraphics[width=0.66\textwidth]{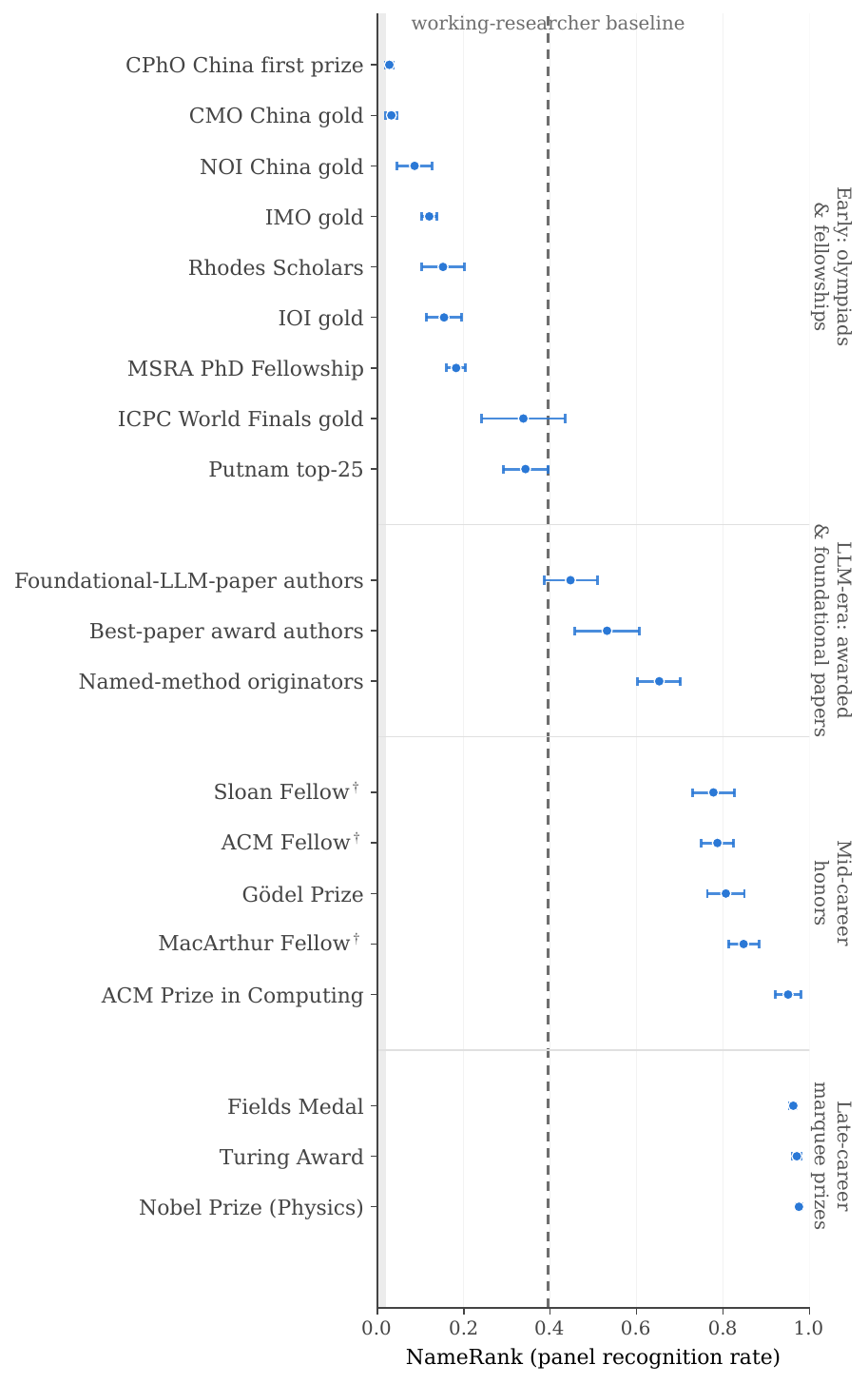}
    \caption{\textbf{Recognition of credentials and awards by career stage.} All credential and award cohorts on the recognition axis, with the working-researcher baseline (dashed) and the synthetic-null floor (hatched).}
    \label{fig:career-arc}
\end{figure}

If the mechanism is named-artifact attachment, then a credential that \emph{ships with a named artifact} should propagate. The LLM-era publication credentials confirm it directly (Figure~\ref{fig:career-arc}): authors of best-paper-awarded papers at top venues score $0.53$ and named-method originators $0.65$---both above the working-researcher baseline, because the award or method carries a named, heavily-indexed artifact into the corpus alongside the person, and rising in exactly that order as the artifact attaches more directly to the individual. The marquee prizes sit higher still, at the panel ceiling (Turing, Fields, Nobel laureates all above $0.95$). Two reading notes temper the marquee tier. First, at that ceiling the prize marks fame as much as it made it: an individual laureate is a famous person the judge recognizes directly, and because recognition saturates there is no room for a within-award time trend to read a lifetime clock off the marquee tier (that clock is visible only in the \emph{unsaturated} cohorts; Appendix~\ref{app:awards}). Second, the sampled mid-career fellowship cohorts are upper bounds---they include only recipients prominent enough to have an encyclopedia article. The clean comparison is among complete-roster cohorts on identical rosters, and along that comparison the arc rises steadily.

The groups of Figure~\ref{fig:career-arc}---olympiad, LLM-era publication, mid-career honor, marquee prize---trace it: recognition rises with the density of named artifacts a credential carries, not with the intellectual selectivity of the filter that awarded it. This is the paper's organizing result.

\section{Why Named Artifacts Drive Recognition}
\label{sec:results-mechanism}

If recognition is paid to named artifacts, the artifact should sometimes out-rank the person who made it. It does, and the direction of the effect---plus a control that comes back null---pins the mechanism to corpus attribution rather than anything a prompt can manufacture.

\subsection{Artifacts usually score higher than their creators}

For verified (creator, artifact) pairs we measure both members independently (Figure~\ref{fig:inversion}a). In $7$ of the $10$ resolvable pairs, the artifact out-ranks its creator, and for independent creators of a single artifact the inversion is uniform: Jiayi Weng ($0.22$) below Tianshou ($0.78$), Harrison Chase below LangChain, Aman Sanger below Cursor, Simon Willison below Datasette, Tri Dao below FlashAttention, Aravind Srinivas below Perplexity. The three exceptions---Demis Hassabis, Andrej Karpathy, Mira Murati---are senior leaders who accumulated independent recognition before, or beyond, the artifact in question; Karpathy is famous for too many things for nanoGPT to carry him. The pattern extends beyond these hand-verified pairs into the LLM-methods landscape, where the same inversion recurs: ReAct ($0.81$) out-ranks Shunyu Yao ($0.67$), who introduced it, and DPO ($0.92$) out-ranks Rafael Rafailov ($0.61$); among named methods prominent enough to appear as standalone entities, the method beats its originator in almost every case (Figure~\ref{fig:inversion}b). The named-method-originator cohort as a whole ($0.65$) still sits far above the researcher baseline---each member's method drags their name into the corpus---but the method's own name travels further than the person's.

The direction is an asymmetry in the corpus itself, not a symmetric association. Measuring how often a probe about one member names the other, the common regime is \emph{famous artifact, anonymous creator}: a probe about the creator reliably surfaces the artifact, but a probe about the artifact frequently omits the creator. Open-source and academic norms produce the opposite, balanced pattern (creator and artifact co-occur in both directions), and the inverse case---Karpathy and nanoGPT---shows a probe about the artifact reliably surfacing the creator while the reverse fails (per-pair breakdowns in Appendix~\ref{app:asymmetry}). This attribution asymmetry is exactly what makes the artifact out-rank its maker: the corpus discusses the tool far more often than the person behind it.

\subsection{Naming an artifact does not raise its creator's recognition}
\label{sec:results-mediation-causal}

A natural causal question: if we \emph{tell} the model the creator made the artifact, does recognition of the creator rise? We re-probed all $11$ pairs under a controlled intervention---configuration~A gives a role-only context, configuration~B appends one clause naming the artifact---and scored both with the recognition judge, which credits only facts \emph{beyond} the context. The lift is essentially zero (mean $-0.01$; positive only for the two least-known creators, Jiayi Weng and Tri Dao). This is the informative null: an earlier scoring rule that credited coverage showed an apparent lift, but that lift was the model echoing the injected artifact name back: once the judge stops rewarding the echo, naming the artifact adds nothing. Recognition of the creator is written into the corpus or it is not; a prompt cannot manufacture it. The inversion is thus an observational property of how attribution is distributed in indexable text, which is exactly what the mechanism predicts.

\subsection{Widely used tools whose names do not spread score low}
\label{sec:results-cases}

The mechanism also says what recognition is \emph{not}. Two Chinese-language sites with enormous user bases make the point: tuixue.online, which monitored US visa-appointment availability for a peak audience near one million, scores $0.39$, and icourse.club, a widely-used university course-review site, scores just $0.14$---both below nanoGPT and vLLM, whose audiences are orders of magnitude smaller (Figure~\ref{fig:inversion} caption; full comparison in Appendix~\ref{app:asymmetry}). Each spread \emph{functionally}---users passed around what the site \emph{does} (``the site has visa slots,'' ``look up a course'') rather than its name---with no creator name attached in indexable, English-readable text; for tuixue the gap is symmetric across English and Chinese prompts, so it is not a language artifact. Recognition measures indexed name presence, not utility: the two correlate wherever creators attach names to their work and dissociate for utility-driven tools that spread anonymously.

\begin{figure}[!htbp]
    \centering
    \includegraphics[width=\textwidth]{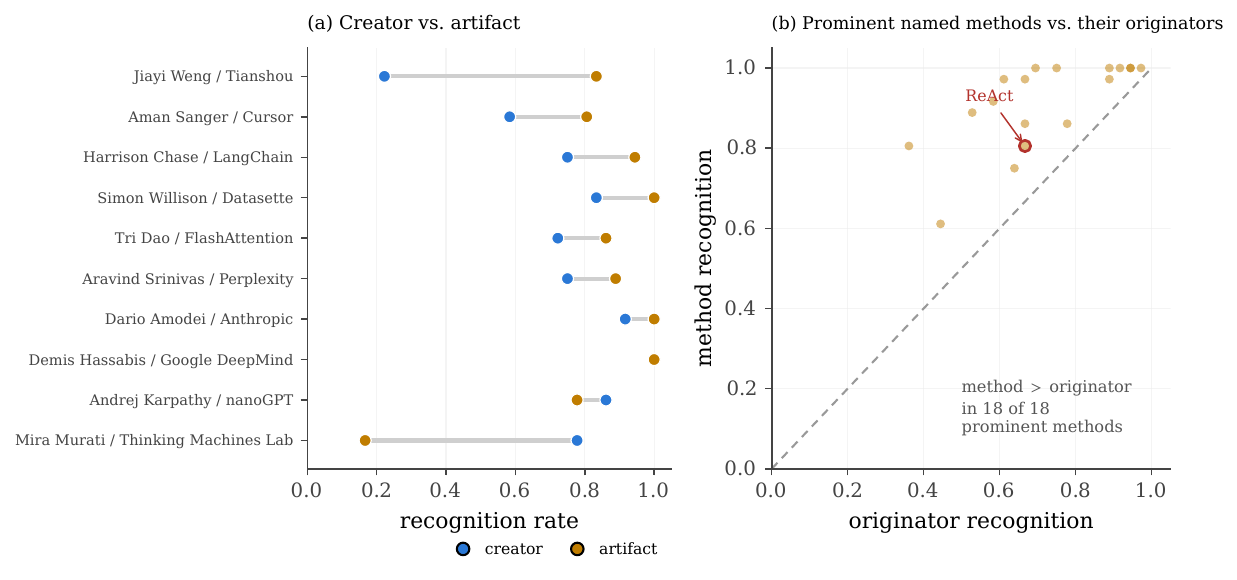}
    \caption{\textbf{Artifacts usually score higher than their creators.} \textbf{(a)}~Creator vs.\ artifact recognition for the $10$ resolvable pairs (blue creator, amber artifact), sorted by artifact$-$creator delta. \textbf{(b)}~Named methods vs.\ their originators.}
    \label{fig:inversion}
\end{figure}

\section{What Predicts Recognition}
\label{sec:results-external}

Turning from named artifacts to measurable correlates: what, among the quantities institutions actually track, predicts whether a name has propagated?

\subsection{No bibliometric predicts recognition well}

\begin{figure}[!htbp]
    \centering
    \includegraphics[width=\textwidth]{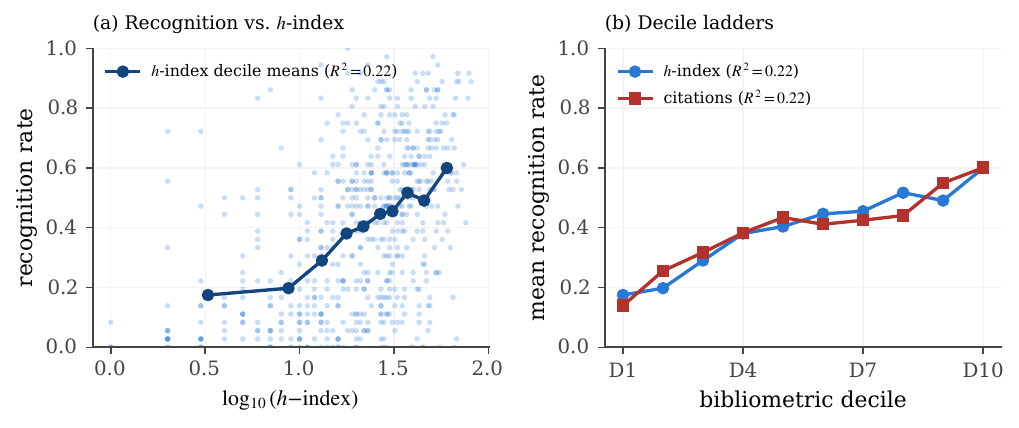}
    \caption{\textbf{Recognition vs.\ bibliometrics.} \textbf{(a)}~Recognition vs.\ $\log_{10}(h\text{-index})$ for OpenAlex researchers, with decile means. \textbf{(b)}~Decile ladders for $h$-index and raw citations.}
    \label{fig:hindex}
\end{figure}

On the working-researcher cohort ($603$ researchers with OpenAlex bibliometrics), recognition rises with bibliometric productivity (Figure~\ref{fig:hindex}), but no single measure dominates: $\log_{10}(h\text{-index})$, raw citations, and attribution-weighted variants all explain around a fifth of the variance ($R^2 \approx 0.22$; Appendix~\ref{app:bibliometric}), and the differences among them are within noise. The headline is the ceiling, not the ranking: even the best bibliometric leaves roughly three-quarters of recognition variance unexplained. That residual is exactly what this instrument exists to measure---the part of recognition that flows not through the publication system but through the named-artifact and corpus-density channels of the surrounding sections. A bibliometric can rank researchers within the academy; it cannot stand in for whether a name has propagated.

\subsection{Recognition rises with institution and country web presence}
\label{sec:results-geography}

\begin{figure}[!htbp]
    \centering
    \includegraphics[width=\textwidth]{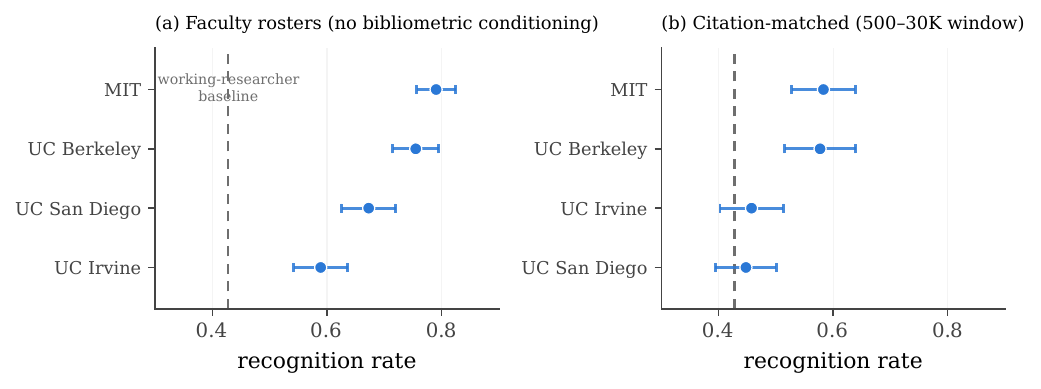}
    \caption{\textbf{Recognition by institution, two analyses.} \textbf{(a)}~CS-faculty recognition by university (CSRankings rosters). \textbf{(b)}~The same at a matched citation window. Dashed line: working-researcher baseline.}
    \label{fig:universities}
\end{figure}

\begin{figure}[!htbp]
    \centering
    \includegraphics[width=0.72\textwidth]{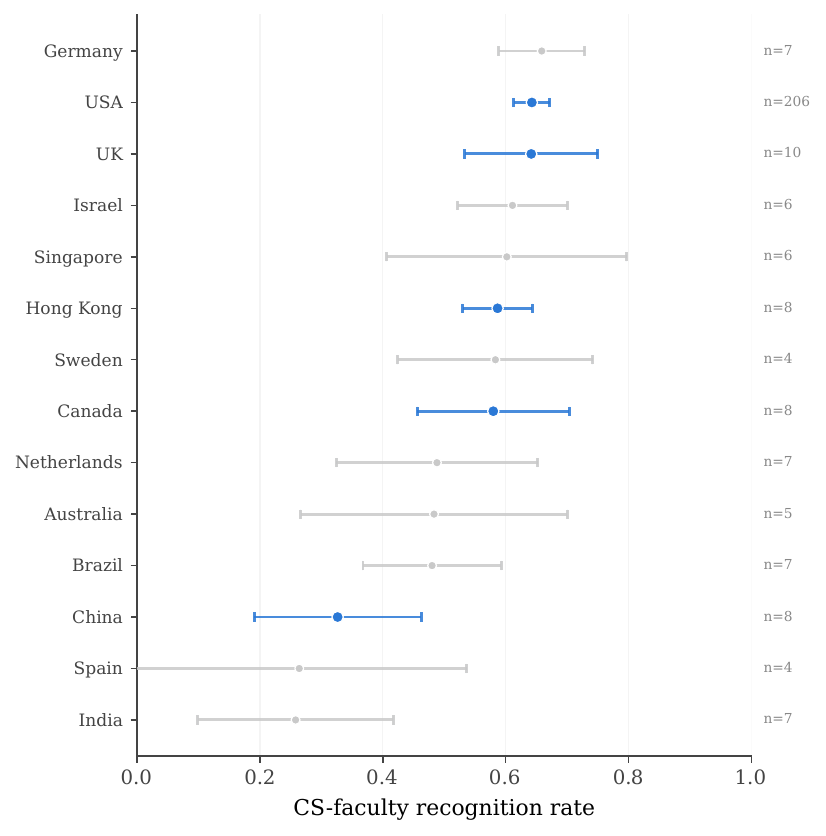}
    \caption{\textbf{CS-faculty recognition by country} (dot, $95\%$ CI; faded rows $n<8$, suggestive only).}
    \label{fig:country}
\end{figure}

Recognition varies with where a researcher works, and---crucially---not only through productivity. Among CS faculty, top-corpus-density institutions lead: an MIT faculty member scores $0.79$, Berkeley $0.76$, against lower-corpus-density peers (Figure~\ref{fig:universities}a). The clean test holds citations fixed: restricting to a matched citation window (Figure~\ref{fig:universities}b), MIT and Berkeley faculty still out-recognize UCSD and Irvine faculty at the \emph{same} citation level---so the gradient is corpus density, not research output. Aggregated by country (Figure~\ref{fig:country}), the firm, well-sampled contrast is the USA ($0.64$, $n=206$) above China ($0.33$) and India ($0.26$), whose confidence intervals do not overlap the USA's; small-sample leaders are read as suggestive only. The mechanism is English-corpus density: identical research output propagates through a denser English-language ecosystem of social media, newsletters, podcasts, and tutorials for a US-based researcher than for a peer whose output travels mainly through the papers themselves.

\subsection{For news events, recognition tracks peak attention}
\label{sec:results-events}

\begin{figure}[!htbp]
    \centering
    \includegraphics[width=\textwidth]{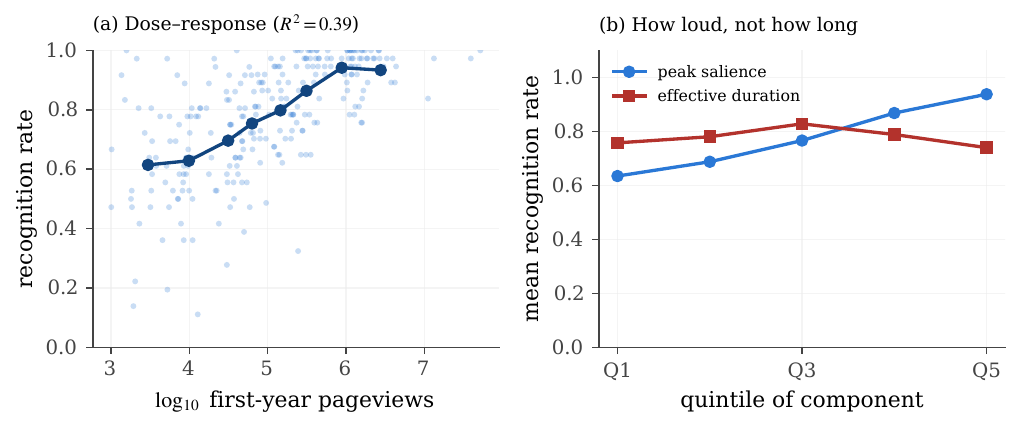}
    \caption{\textbf{Recognition versus news-event attention} ($258$ news events, $2021$--$2023$). \textbf{(a)}~Recognition vs.\ $\log_{10}$ first-year pageviews, with decile means. \textbf{(b)}~Quintile ladders for the two components of attention, peak salience and effective duration.}
    \label{fig:events}
\end{figure}

Every predictor so far proxies corpus exposure after the fact. News events invert the problem: the attention the world paid an event is itself a public, time-stamped record, so here the exposure side of the corpus-propagation account can be measured directly rather than inferred. We built a calibration cohort of $258$ news events from $2021$--$2023$, each carrying its English-Wikipedia daily pageview series, and probed them with the unchanged pipeline (Appendix~\ref{app:events}). Recognition rises log-linearly in total first-year attention. The decomposition is the sharp result: total attention factors into peak $\times$ duration, and \emph{peak} dominates (standardized coefficient $+0.13$ against $+0.04$ on duration; duration adds a marginal $R^2$ of only $0.045$, against $0.34$ explained by peak alone). How \emph{loud} a story got, not how \emph{long} it lasted, drives recognition---because a story's peak marks how many name-bearing documents were minted at climax, while a long tail of readership mints little additional text. Recognition follows document production, not audience retention; it is the temporal echo of the named-artifact mechanism, and it runs opposite to the $h$-index intuition that persistence should matter (Figure~\ref{fig:events}).

\section{Robustness and Validation}
\label{sec:results-validation}

The findings above rest on three design choices---an LLM judge over embedding similarity, a binary recognition verdict over a graded score, and open-ended over closed-factoid probes---and on the headline numbers not being artifacts of wording, model vintage, judge family, or gold-answer construction. Appendices~\ref{app:scoring}--\ref{app:confounds} give the full evidence, summarized threat-by-threat as a validation battery in Table~\ref{tab:threats}; the load-bearing checks are expanded here, and two checks that stand as their own experiments---self-report and prompt language---get their own subsections.

Three of these checks carry most of the weight. First, \emph{the scoring rule does the essential work}: the entity factor explains roughly six times as much variance as the model factor, and a fresh replacement panel reproduces the event ordering, so the score measures the entity, not the fleet; the recognition verdict is what catches the models that never refuse and confidently fabricate---responses an embedding score would credit as recognition (Appendix~\ref{app:scoring}). Second, \emph{the silent zone is genuine obscurity, not corpus timing}: on two author cohorts published at the same time, crossing a model's training cutoff raises the more recent cohort \emph{less} than it raises stable controls (matched difference-in-differences $-0.24$ and $-0.17$), even as a ${\sim}0.1$--$0.3$/yr capability drift accumulates across model vintages---so absolute levels are comparable only within a single run, but the silence itself is intrinsic (Appendix~\ref{app:cutoff}), as the literature on how transformers store facts predicts~\citep{geva2021transformer, petroni2019language, kandpal2023, overshadow2025}. Third, \emph{the floor and confound checks hold}: adjusting for the synthetic-null floor leaves every credential below the working-researcher baseline, and dedicated ablations reject both Wikipedia presence and web-attention rank as alternative explanations (Appendices~\ref{app:synthetic}, \ref{app:confounds}). Orderings survive paraphrase and judge family (a three-family re-judge reproduces the cohort ranking under Gemini, Claude, and a uniformly stricter GPT-5.1); only absolute levels shift, so NameRank values are read as within-run relative gaps. Two further checks were run as standalone experiments and get their own treatment below: whether a model can simply be \emph{asked} how well it recognizes an entity, and whether the score survives a change of probe language.

\subsection{Asking a model what it knows does not replace measurement}
\label{sec:results-selfreport}

\begin{figure}[H]
    \centering
    \includegraphics[width=0.9\textwidth]{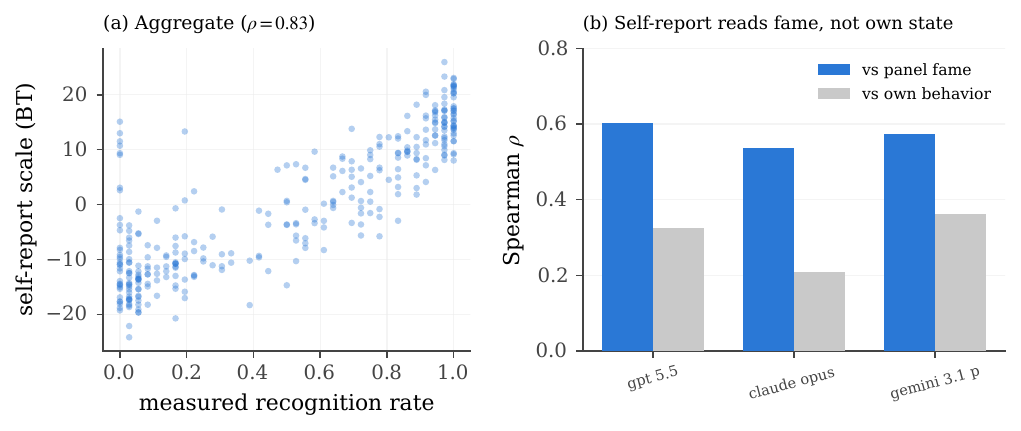}
    \caption{\textbf{Model self-report reflects overall fame, not the model's own knowledge.} \textbf{(a)}~Bradley--Terry self-report scale vs.\ measured panel recognition. \textbf{(b)}~Per model, self-report vs.\ cross-panel fame and vs.\ the model's own behavioral record.}
    \label{fig:selfreport}
\end{figure}

Could a model simply be asked how well-known an entity is? We put this to a pairwise test: models rank which of two entities they recognize better, and we fit a Bradley--Terry scale from the comparisons (Appendix~\ref{app:selfreport}). In aggregate the self-report scale recovers the panel recognition ordering reasonably well (Spearman $\rho \approx 0.83$; Figure~\ref{fig:selfreport}a)---but the recovery is borrowed, not introspective. Two diagnostics show why. First, each model's self-ranking tracks how \emph{widely} a name is known across the panel ($\rho \approx 0.6$) far better than its own behavioral record on the entities it demonstrably knows ($\rho \approx 0$; Figure~\ref{fig:selfreport}b). Second, the three vendors' self-report scales agree with \emph{each other} ($\rho \approx 0.90$) more than any agrees with the ground-truth panel recognition---they are reading a shared corpus prior, not their own weights \citep{kadavath2022language, yin2023selfaware}. In one direction the claims are still trustworthy: opposite a name the model genuinely knows, the fictional side of a trap pair was almost never preferred. Self-report errs by reciting consensus fame, not by fabrication---usable as a prior, insufficient as measurement.

\subsection{Recognition is the same for English and Chinese prompts}
\label{sec:results-language}

Re-probing a $240$-entity subset with a Chinese-translated template moves no cohort's recognition rate by more than $0.04$ (Appendix~\ref{app:cross-lang}). This is the expected behavior of a recognition verdict: because the judge credits only a specific non-guessable fact, and whether the model stores that fact is a property of its weights rather than of the language it is queried in, the wording of the probe does not change what the model knows. An earlier graded-coverage version of this sub-run showed a sizeable Chinese-prompt lift for Chinese-named cohorts, but that was language-matched \emph{verbosity}---a Chinese prompt draws a longer Chinese answer that accrues more coverage points---not extra knowledge; the recognition verdict removes it. A residual nudge (at most ${\sim}0.14$) survives only for low-recognition Chinese-named credential holders, where a Chinese prompt occasionally tips a borderline model over the recognition threshold. NameRank is therefore comparable across prompt languages; all headline results use English prompts.

\section{Discussion}
\label{sec:discussion}

Three questions organize what the findings imply beyond the measurements themselves: what \emph{clock} the score reads, what it says to anyone who wants their own name to propagate, and what structural inequalities it exposes. We then close with limitations and a set of falsifiable predictions for future re-runs.

\subsection{Does NameRank Measure Lifetime or Current Recognition?}
\label{sec:discussion-clock}

A recognition score could measure two different quantities: \emph{cumulative} presence---everything a name has deposited in the corpus over a career---or \emph{current-term} salience, what the world is paying attention to now. Three findings established independently above jointly place NameRank in the first category. Within IMO gold medalists, medal year is essentially uncorrelated with the score (Section~\ref{sec:results-credentials}): recognition neither fades for older medalists nor spikes for recent ones. The marquee awards show the complementary shape: recognition does not decay with time since the prize---decades-old Nobel, Turing, and Fields laureates remain saturated rather than fading (Appendix~\ref{app:awards})---which rules out a \emph{flow} reading, under which old laureates would have faded, even though the saturated marquee tier cannot itself trace the accrual slope (that is left to the unsaturated cohorts; Appendix~\ref{app:awards}). And on news events, where attention is directly recorded, the score loads on \emph{peak} salience rather than persistence (Section~\ref{sec:results-events}): what matters is how many name-bearing documents were minted, not how long the audience stayed. Read together: NameRank is a cumulative ledger of name-bearing text, captured through the current model fleet---a lifetime measure weighted by documentation density, not a gauge of current attention. Current-term change is accessible only longitudinally, as within-run relative gaps across successive fleets (Section~\ref{sec:predictions}).

\subsection{What Raises Recognition, and Whether Virality Helps}
\label{sec:discussion-implications}

As practical guidance, the findings converge on one strategy: recognition accrues to \emph{distinctively-named, indexable production accumulated over a career}. Volume of name-bearing works leads citation weight ($h$-index over raw citations, Section~\ref{sec:results-external}); one clearly-named artifact outperforms many papers (Section~\ref{sec:results-mechanism}); and the credentials that propagate on their own are exactly those that carry a named artifact---best-paper awards, named methods, marquee prizes backed by a career of production---while a bare olympiad medal does not (Section~\ref{sec:results-awards}). Because the ledger is cumulative (Section~\ref{sec:discussion-clock}), the effective strategy is long-term and structural rather than about any single achievement: keep depositing documents that carry a distinctive, low-collision name into English-indexed text. One caveat is attribution-direction: recognition accrues to an artifact's name first and to its creator only through the corpus's attribution chain (Section~\ref{sec:results-mechanism}), so a tool buried under generic corporate branding builds no personal recognition.

This also bounds what public attention buys. Going viral helps only insofar as the spike is \emph{captured as durable, name-bearing documents}: on news events the score loads on peak salience over persistence (Section~\ref{sec:results-events}), so a burst that mints many name-attached records converts while ephemeral engagement that leaves none does not. The conversion is category-dependent: within celebrity cohorts pageviews track recognition closely, but for technical creators attention \emph{flow} and corpus \emph{stock} largely decouple (Appendix~\ref{app:confounds}; a low-traffic but corpus-saturated library can outscore a high-traffic recent model). Attention is thus a means, valuable only when it deposits distinctively-named text into the corpus---never an end in itself.

\subsection{Consequences: Unequal Web Presence and Inherited Bias}
\label{sec:discussion-bias}

The corpus-density gradient of Section~\ref{sec:results-geography}---top-density institutions above their peers even at matched citations, recurring at country scale with the USA-over-China-over-India ordering at its core---is a structural inequality in name propagation \emph{for a given amount of research output}. It reflects training-pipeline indexing decisions rather than research quality.

NameRank also inherits the demographic biases of its judge and probed models and measures them faithfully rather than correcting them: beyond the ${\sim}2\times$ East-Asian-surname attenuation documented by \citet{li2026ikp}, man-coded names outscore woman-coded names by a significant $+0.085$ among CS faculty (directional-only in the smaller researcher cohorts; Appendix~\ref{app:gender}). We restrict the reading to the research cohorts---the larger cross-profession gaps (actors $+0.21$, athletes $+0.27$) reflect real fame differences in the sample, not a NameRank artifact---and report the gap as an inherited corpus-and-judge property, not a calibration target, that downstream uses must account for.

\subsection{Limitations}

The material caveats, each with its direction and mitigation: absolute levels are wording-conditional while rankings are robust; the run uses a single judge family, checked against others; the marquee-award tier measures the fame the prize both marks and made, and the sampled fellowship cohorts are upper bounds; only within-run \emph{relative} gaps are comparable across model fleets; and paper and competition cohorts carry a small residual synthetic-null floor that is subtracted where relevant.

\subsection{Registered Predictions for Future Re-Runs}
\label{sec:predictions}

NameRank is a snapshot, so a re-run each release cycle yields a longitudinal trajectory. Because absolute levels drift across model fleets, we state predictions as within-run \emph{relative} gaps (so the drift cancels) and register four falsifiable ones for the next fleet: (i)~the IMO-gold-vs-working-researcher gap (currently $\approx -0.28$) widens further within 12 months, as named-artifact-attached careers accumulate text while credential-only cohorts do not; (ii)~the median artifact-minus-creator delta on independent-creator pairs grows within 18 months; (iii)~the top-vs-lower-corpus-density institution gap narrows but does not close within 24 months as more non-English training data enters frontier models; (iv)~on a re-run of the news-event cohort (Section~\ref{sec:results-events})---the cleanest of the four, since its exposure side is frozen in the pageview archive---the recognition gap between older and newer events at matched attention shrinks as post-event derivative text accumulates.

\section{Related Work}
\label{sec:related}

\subsection{Relation to the Predecessor Instrument}

The direct predecessor of this work is the recognition analysis in the IKP benchmark \citep[\S5.7]{li2026ikp}, which established the findings this paper builds on: on $345$ CS-researcher probes, $\log(\text{citations})$ explains only ${\sim}35\%$ of recognition variance, with the remainder attributed in a controlled audit to \emph{named-artifact amplification}, \emph{name uniqueness} (common East Asian surnames attenuate recognition by ${\sim}2\times$), and \emph{subfield-ecosystem density}; and on $557$ Wikidata-grounded probes, domain-specific recognition multipliers trace to the structure of web discourse around each fact type rather than to entity prominence. IKP nonetheless treated recognition as a benchmark byproduct. NameRank differs in each dimension that matters for an impact metric: a $[0,1]$ recognition rate over a $36$-model panel (vs.\ a six-landmark tier ladder calibrated for parameter estimation); thousands of entities across dozens of cohorts spanning industry, OSS, and mid-tier figures (vs.\ $345$ researchers and $557$ Wikidata probes); open-ended probes scored by a per-model recognition verdict (vs.\ closed-factoid binary verdicts); $h$-index rather than raw citations as the validated bibliometric correlate; and cross-model variance retained as signal rather than absorbed into a tier label.

\subsection{Scientometric Measures of Researcher Impact}

The classical instruments for researcher impact---citation count, $h$-index~\citep{hirsch2005index}, journal impact factor---measure intellectual reach within the academic publication system, and are increasingly understood as failing under pressures relevant here: authorship-list inflation and uneven attribution~\citep{lariviere2019}, the \emph{hyperauthorship} problem of pricing contributions to thousand-author papers~\citep{cronin2001hyperauthorship} (a GPT-4 co-author's citation count is mechanically large but says nothing about name-recognition), prestige-hierarchy biases spanning an order of magnitude across institutional tiers~\citep{wapman2022quantifying}, and within-researcher variance that motivates decompositions like the $Q$ parameter~\citep{sinatra2016quantifying}.

None of these instruments cross the academic boundary cleanly. Citation count does not measure an industry executive or a startup founder, and $h$-index does not measure the creator of a widely-used open-source tool (Section~\ref{sec:results-headline}). Bespoke domain metrics (Google PageRank for websites, GitHub stars for OSS, Spotify monthly listeners for musicians, IMDb scores for film) do not translate into one another. Collective-attention metrics do---search volume, Google Trends, and Wikipedia pageviews place any entity on one axis, and pageviews in particular are an established salience proxy~\citep{mestyan2013early,ripberger2011capturing}---but they count queries and article traffic rather than verified knowledge, and they are empirically almost unrelated to NameRank in the zone that matters: undefined for the majority of entities that lack an article, and only weakly correlated with recognition within the technical and research cohorts that carry the findings (Appendix~\ref{app:confounds}). We use attention records as calibration input on the event cohort (Section~\ref{sec:results-events}), not as the quantity we measure: NameRank reads the \emph{outcome} of the training pipeline---what deployed models can correctly state---rather than the exposure that feeds it. The cross-domain measurement of eminence has a long prior tradition in \emph{historiometry}~\citep{simonton1999}, which quantifies the reputational standing of eminent individuals from archival traces; NameRank can be read as a historiometric instrument whose archive is the frontier-model training corpus rather than biographical dictionaries and citation indices. For impact assessment that spans academia, industry, OSS, and the long tail of solo operators, a cross-domain instrument is needed; NameRank is the first we are aware of that reads this archive directly.

\subsection{LLM-as-Judge for Quality and Impact}

\citet{thelwall2024chatgpt, thelwall2025plos, thelwall2025multistage} use ChatGPT to score paper \emph{quality} against the UK REF rubric; \citet{kousha2025jasist} extend to societal influence; \citet{liu2026llmprediction} forecast citation counts. These differ from NameRank in two respects: they ask the LLM to \emph{judge} quality, where we ask what the model has \emph{absorbed}; and they average within a single model across iterations, where we average across the full $36$-model panel, treating cross-model variance as signal.

A separate line asks whether models \emph{know what they know}: self-assessed confidence is well calibrated on aggregate question answering \citep{kadavath2022language} but degrades at the boundary of a model's own competence \citep{yin2023selfaware}, and entity popularity modulates factual reliability \citep{mallen2023trust}. Section~\ref{sec:results-selfreport} adds an entity-level counterpart: pairwise self-reported recognition, aggregated with the Bradley--Terry machinery familiar from Chatbot Arena \citep{chiang2024chatbot}, tracks the corpus salience shared across vendors rather than the reporting model's own verified knowledge.

The known LLM bias most relevant to our setting is the \emph{Matthew effect}~\citep{merton1968matthew} as it propagates through LLM training corpora: frontier models recall already-highly-cited work disproportionately, inflating the rich-get-richer dynamic familiar from the bibliometrics literature. We inherit this bias and treat it as a known property of what we measure---NameRank measures \emph{what frontier models have absorbed}, which is itself shaped by the Matthew dynamic. The bias is part of what we measure, not a flaw to be corrected.

\subsection{Credentials and Long-Run Outcomes}

A longitudinal literature finds that early mathematical talent and olympiad medals predict---and causally raise---later \emph{academic} productivity: the 50-year SMPY study~\citep{smpy2020}, a medal-cutoff natural experiment~\citep{agarwalGaule2020}, and work documenting substantial within-elite heterogeneity~\citep{guellich2025}. Our credential-treadmill finding does not contradict this literature; it changes the dependent variable from academic production to LLM-mediated name recognition and finds that the credential per se carries little of it (Section~\ref{sec:results-credentials}). The credential still signals ability and motivation; only its role as a name-propagation channel has shrunk---in the language of signaling theory~\citep{spence1973}, the LLM-corpus channel transmits not the credential itself but the named, indexable production that may follow it.

\section{Conclusion}
\label{sec:conclusion}

NameRank measures the recognition-variance residual that \citet{li2026ikp} characterized but did not quantify: a cross-domain, hallucination-resistant recognition rate over a $36$-model panel, whose findings each subvert a default expectation---olympiad credentials no longer propagate names, named tools out-rank their makers, and a model asked to rank its own recognition returns the shared corpus prior rather than its own knowledge. The broader claim is that human curiosity about people and artifacts is now re-mediated by a few frontier models, and whether a name has propagated into their training corpora is itself a measurable quantity---usable as a cross-domain impact metric, a longitudinal tracker of how discourse reaches training pipelines, and a normative indicator for anyone who wants their name to survive the transition. The instrument measures the post-bibliometric impact channel; what that channel rewards is not the credential you were awarded but the named, indexable artifact you built and shipped.

\section*{Acknowledgements}

This paper was produced using Pine Copilot's voice-directed \emph{whisper coding} workflow~\citep{pineai2026whispercoding}, in which the authors specify, discuss, and review the work by voice while a coding agent---Claude Code with Claude Opus 4.8 and Claude Fable 5---carries out the planning, coding, experiments, and paper writing.

\bibliographystyle{plainnat}
\bibliography{references}

\clearpage
\appendix
\input{appendix}

\end{document}

%% file: appendix.tex
\begin{center}
\small\emph{The appendices are organized in three blocks. Appendix~\ref{app:spec} completes the instrument specification of Section~\ref{sec:method}. Appendices~\ref{app:awards}--\ref{app:selfreport} give the full tables and case detail behind the findings of Sections~\ref{sec:results-headline}--\ref{sec:results-validation}, roughly in the order they appear. Appendices~\ref{app:scoring}--\ref{app:confounds} give the evidence behind the validation battery of Section~\ref{sec:results-validation}, one appendix per threat; Appendix~\ref{app:gender} covers the inherited-bias discussion.}
\end{center}

\section{Probe, Panel, and Cohort Specification}
\label{app:spec}

\textbf{Why disambiguation does not contaminate scoring.} The probe context names role, affiliation, and field---never specific contributions, papers, dates, or named artifacts---and the recognition judge credits only facts that go \emph{beyond} what the context supplies (Section~\ref{sec:method-judge}). A model that recognizes the entity states specific facts the context did not give; a model that does not produces generic statements about the role, which earn no recognition. The sharpest empirical check comes from the GPT-5 system-card author cohort, whose disambiguating context is identical across all $80$ entities and which scores at the synthetic-null floor: the panel withholds recognition rather than manufacturing a generic in-context biography. A dedicated context-ablation (Appendix~\ref{app:confounds}) confirms the same conclusion.

\begin{table}[!htbp]
\centering
\footnotesize
\caption{The $36$-model panel, by vendor class. Thinking-mode variants are used wherever available; the panel spans frontier-reasoning to small (sub-$30$B) classes.}
\label{tab:panel}
\begin{tabularx}{\textwidth}{XX}
\toprule
\textbf{Western (n=20)} & \textbf{Chinese (n=16)} \\
\midrule
GPT-5.3, GPT-5.4, GPT-5.5-think, GPT-OSS-20b-think & DeepSeek-V3.2-think, V4-flash-think, V4-pro-think \\
Claude Opus 4.6-think, Sonnet 4.6-think, Fable 5-think & Qwen3-8b-think, 32b-think, 235b-a22b-think, 3.5-397b-a17b-think, 3.7-Max-think \\
Gemini 2.5-pro-think, 3-flash-think, 3.1-pro & GLM-4.7-think, 5.1-think, 5.2-think \\
Llama-3.1-8b, 3.3-70b, 4-Maverick & Kimi-K2.6-think, K2.7-code-think \\
Mistral-Small-24b, Phi-4 & MiniMax-M2.7-think, M3-think \\
Gemma-3-4b, 3-12b, 4-31b & Step-3.7-Flash-think \\
Nemotron-3-Ultra-think, Nano-30b-think & \\
\bottomrule
\end{tabularx}
\end{table}

\begin{table}[!htbp]
\centering
\small
\caption{Core-cohort entities by top-level group (main dataset; $4{,}685$ real entities plus $45$ synthetic-null controls). Extensions are specified in their own sections; full per-cohort breakdown in the \repodoc{cohorts.md}{released cohort specification}.}
\label{tab:cohort-groups}
\begin{tabularx}{\textwidth}{lrX}
\toprule
Group & $n$ & Representative cohorts \\
\midrule
Credentialed individuals & 801 & IMO/IOI/CMO/NOI/CPhO medalists, Putnam, ICPC, Rhodes, MSRA Fellowship \\
Working academics & 1{,}210 & CS faculty; OpenAlex/IKP researchers with $500$--$30{,}000$ citations \\
Industry figures & 122 & founders, VCs, YC companies, AI-policy officials \\
Named artifacts & 1{,}115 & foundation models, OSS projects, papers, benchmarks, named methods, datasets \\
Mid-tier cross-profession & 1{,}214 & writers, athletes, actors, chefs, journalists, podcasts, books \\
Diagnostic reference set & 37 & hand-curated public figures, engineers, and artifacts \\
Special cohorts & 186 & GPT-5 system-card authors, DeepSeek-V3 authors, conferences, awards \\
\bottomrule
\end{tabularx}
\end{table}


\section{Mid/Late-Career Awards as Recognition Predictors}
\label{app:awards}

This appendix gives the construction and full tables behind Section~\ref{sec:results-awards} and Figure~\ref{fig:career-arc}.

\textbf{Cohorts.} Laureate rosters were pulled from Wikidata (award-received statements, with the award-date qualifier supplying the year): Turing Award ($77$ human laureates with an English label), Fields Medal ($64$), Nobel Prize in Physics restricted to $2000$--$2023$ ($64$), G\"odel Prize ($80$), ACM Prize in Computing ($18$), and, sampled to $60$ each from their $2000$--$2023$ Wikidata-listed recipients, the MacArthur Fellowship and ACM Fellow honors; the Sloan Research Fellowship ($49$ listed for $2000$--$2023$) is the early-career anchor. The five prize cohorts (Turing, Fields, Nobel, G\"odel, ACM Prize) are complete rosters by construction---every laureate has a Wikidata item. The three fellowship/honor cohorts (MacArthur, ACM Fellow, Sloan) are sampled from Wikidata's coverage, which omits lower-profile recipients, so their means are \emph{upper bounds} and are read as such.

\textbf{Panel, golds, and contexts.} These cohorts are probed on the same $36$-model panel and scored by the same recognition judge as every other cohort. Contexts follow the credential convention---award name and year only, never contributions (e.g., \emph{``a mathematician who received the Fields Medal in 2014''}). Gold answers are the laureate's en-Wikipedia lead section where available, otherwise a factual award-and-affiliation record, so recognition is credited only for facts the model states beyond the award itself.

\begin{table}[!htbp]
\centering
\small
\caption{The award ladder under the recognition metric. Recognition is the panel fraction; \emph{vs.\ base} is relative to the working-researcher baseline (recognition $0.46$ on this run's matched $n{=}60$ subsample; IMO control $0.19$ on $n{=}40$), on the same panel and judge. Sampled mid-career honors are upper bounds.}
\label{tab:award-ladder}
\begin{tabular}{llrrr}
\toprule
Award & Stage & $n$ & Recognition & vs.\ base \\
\midrule
Nobel Prize in Physics ($2000$--$23$) & late & $64$ & 0.98 & $+0.52$ \\
Turing Award & late & $77$ & 0.97 & $+0.51$ \\
Fields Medal & late & $64$ & 0.96 & $+0.50$ \\
ACM Prize in Computing & mid & $18$ & 0.95 & $+0.49$ \\
MacArthur Fellowship$^\ddagger$ & mid & $60$ & 0.85 & $+0.39$ \\
G\"odel Prize & mid & $80$ & 0.81 & $+0.35$ \\
ACM Fellow$^\ddagger$ & mid & $60$ & 0.79 & $+0.33$ \\
Sloan Research Fellowship$^\ddagger$ & early & $49$ & 0.78 & $+0.32$ \\
Named-method originators & mid & $84$ & 0.65 & $+0.19$ \\
Best-paper award authors & mid & $65$ & 0.53 & $+0.07$ \\
Foundational-paper authors & mid & $85$ & 0.45 & $+0.00$ \\
\textbf{OpenAlex researchers (baseline)} & --- & $60$ & \textbf{0.46} & --- \\
IMO gold (control) & early cred. & $40$ & 0.19 & $-0.27$ \\
\bottomrule
\end{tabular}

\vspace{2pt}
{\footnotesize $^\ddagger$ Wikidata-coverage sample (omits lower-profile recipients); cohort mean is an upper bound.}
\end{table}

\textbf{The career arc, in one table.} Every cohort with a career of named production behind it clears the working-researcher baseline, and the ordering follows the paper's thesis. Two caveats qualify it. First, the individual laureate cohorts saturate near the ceiling---a Nobel, Turing, or Fields laureate is an individually famous person the judge recognizes directly---so the marquee tier measures fame the prize both marks and made, not the prize in isolation. Second, three fellowship cohorts (MacArthur, ACM Fellow, Sloan; marked $\ddagger$) are sampled from Wikidata, which lists only recipients prominent enough to have an article, so their means are upper bounds and should be read as such. The cleanest comparison is therefore among the \emph{complete-roster} cohorts evaluated on identical rosters: the early olympiad credentials (below baseline), the foundational- and best-paper author cohorts and named-method originators (at-to-above baseline, rising with how directly a named artifact attaches to the person), and the complete-roster prizes (Gödel, ACM Prize, and the marquee three) at the top. Along that clean comparison, recognition rises with named-artifact density, exactly as the mechanism predicts.

\textbf{Saturation, not a lifetime clock.} The marquee prizes cannot themselves reveal the accrual clock, because they sit at the ceiling: a Nobel, Turing, or Fields laureate is recognized by essentially the whole panel ($0.96$--$0.98$), so there is no room for recognition to ``keep rising'' with time since the award. Pooling the three marquee cohorts and standardizing recognition against each award's own mean, the trend against years-since-award is flat (slope ${\approx}0.00$ standardized units/yr; non-monotonic across decade bins), as expected at saturation. The cumulative-ledger reading of Section~\ref{sec:discussion-clock} is therefore carried not by the saturated marquee tier but by the \emph{unsaturated} cohorts, where recognition has room to track accumulated named production: the cohort ladder itself (below), the training-cutoff gradient (Appendix~\ref{app:cutoff}), and the NOI case (Appendix~\ref{app:medal-tiers}), where post-medal careers---not the medal---separate the recognized from the silent.


\section{Medal Tiers: The Complete NOI 2009 Roster}
\label{app:medal-tiers}

The credential ladder above measures \emph{gold} medalists only, so it cannot say whether the medal's grade carries signal. To test this we probed the complete medal roster of a single contest --- NOI 2009: every gold, silver, and bronze medalist ($103$ in all) --- on the same $36$-model panel and recognition judge as every other cohort. The roster comes from the official CCF medal list, cross-validated line-by-line against the community OIerDb database (identical names, schools, and scores); the released files include each medalist's contest score and rank. Probe names carry the Chinese characters alongside the romanization, which --- together with the year-and-school context --- disambiguate medalists whose Pinyin collides with common names.

\textbf{Measurement protocol.} This roster is scored under the recognition judge used throughout the paper (Appendix~\ref{app:scoring}). Each medalist's gold answer is an individual, web-grounded profile of who the person actually is --- their real career and named contributions, retrieved by a search-augmented model and length-normalized to ${\sim}55$ words. The judge is shown the probe context and returns a binary \emph{recognized} verdict: it credits recognition only when the response states a specific, non-guessable fact about the person that is true either per the gold or per the judge's own knowledge, and \emph{not} derivable from the disambiguating context, so a response that merely restates ``a 2009 NOI gold medalist from [school]'' is not recognition. Because the entities are competition participants---for whom a judge has no reliable per-individual memory---the anti-confabulation provision applies: fine-grained minutiae asserted via the judge's own knowledge (other-year participations, exact scores) are not credited, only gold-present facts or major documented achievements. The per-entity score is the fraction of the $36$-model panel that recognizes the person.

\begin{figure}[!htbp]
\centering
\includegraphics[width=0.78\textwidth]{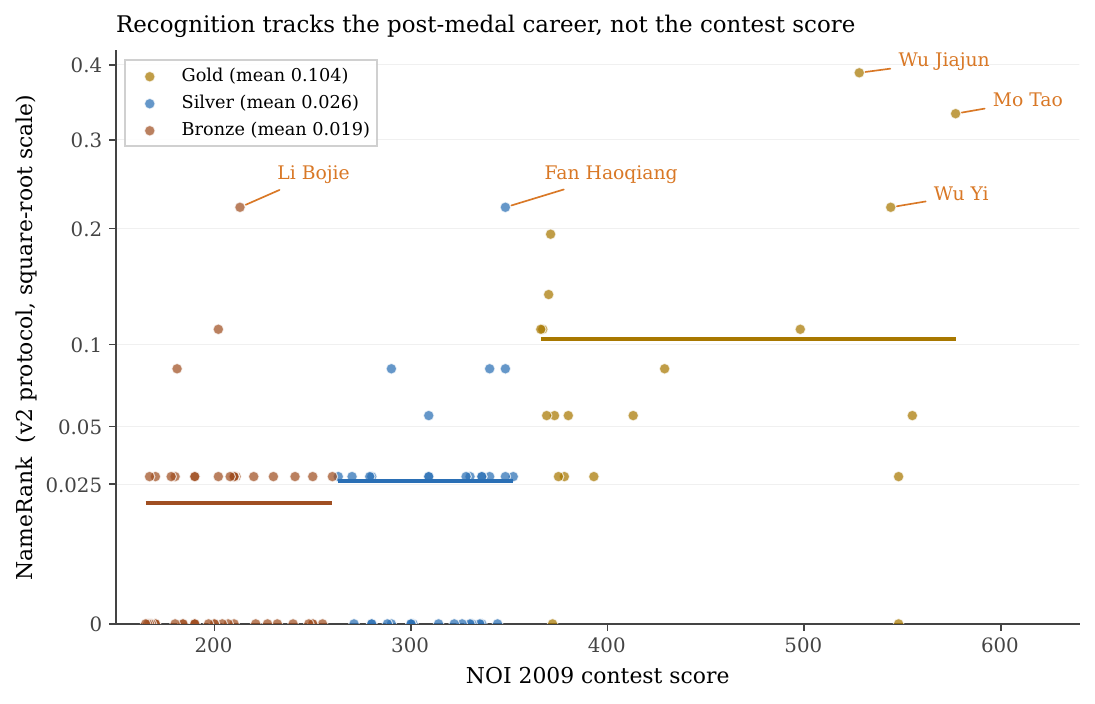}
\caption{NameRank against NOI 2009 contest score for the complete medal roster (square-root $y$-axis), with within-tier means as horizontal lines. Only the five medalists above $0.2$ are labelled.}
\label{fig:medal-tiers}
\end{figure}

\begin{table}[!htbp]
\centering
\small
\caption{NOI 2009 medal tiers under the recognition judge. Score is the panel recognition rate; ``silent'' counts medalists no model recognizes. Synthetic-null floor $0.000$. Kruskal--Wallis $H = 25.9$ ($p < 10^{-5}$); gold vs.\ silver Cliff's $\delta = +0.62$; silver vs.\ bronze indistinguishable ($p = 0.27$).}
\label{tab:medal-tiers}
\begin{tabular}{lrcccc}
\toprule
Tier & $n$ & Mean & Median & Silent ($0$ models) & Recognized ($\geq 1$) \\
\midrule
Gold & 20 & 0.104 & 0.056 & 2/20 & 18/20 \\
Silver & 34 & 0.026 & 0.028 & 16/34 & 18/34 \\
Bronze & 49 & 0.019 & 0.000 & 28/49 & 21/49 \\
\bottomrule
\end{tabular}
\end{table}

\textbf{Most of the population sits near the floor, and that is the finding.} The typical NOI medalist is recognized by only a handful of the $36$ models: the gold-tier median is $0.056$ (about two models), the silver median $0.028$, the bronze median $0$ (Table~\ref{tab:medal-tiers}, Figure~\ref{fig:medal-tiers}). Even in the gold tier, where $18$ of $20$ are recognized by at least one model, the mass is a thin near-floor band with a short recognized tail---a frontier panel stores only a small, documented slice of even this pre-selected population. This is the silent-not-misleading property (Section~\ref{sec:results-headline}) operating at full strength: against a synthetic-null floor of $0.000$, the gold mean of $0.104$ is a real above-floor signal, and the silver/bronze means ($0.026$, $0.019$) sit just above it.

\textbf{The gold/non-gold cliff survives; silver and bronze do not separate.} The gold tier stands clear of the lower tiers (Cliff's $\delta = +0.62$ vs.\ silver, Kruskal--Wallis $p < 10^{-5}$), while silver and bronze are statistically indistinguishable ($\delta = +0.13$, $p = 0.27$). The ${\sim}20$-person national gold roster is the documented one---admissions coverage, competitive-programming lore, team-selection reporting---and it carries by far the most recognition mass (mean $0.104$ against $0.026$ and $0.019$). Contest score \emph{within} a tier does not predict recognition (Figure~\ref{fig:medal-tiers}): the effect is the roster cut, not the marginal point.

\textbf{The recognized names are career-driven --- the treadmill again.} The medalists who rise above the near-floor band are exactly those whose \emph{post-medal} careers produced named, indexed work: the most-recognized medalist overall is a gold who is now a Stanford computer-vision professor (Wu Jiajun, $0.39$); the most-recognized silver matches the gold-tier top on a prominent vision-research career (Fan Haoqiang, $0.22$, later an IOI gold and a Megvii research lead); and the most-recognized bronze ranks near the bottom of the contest's score column yet ties the top silver (Li Bojie, $0.22$, the author of this paper). The medal grade sets the odds of entering a documented cohort; the recognition itself tracks what the person later built.

\textbf{No discontinuity at the medal threshold.} The tier gap could still reflect the gold credential itself---and the guaranteed Tsinghua/Yao-Class admission it triggers, where silver and bronze medalists instead enter other universities by general examination. A boundary comparison rules this out. The gold/silver cutoff sits in a $14$-point gap (lowest gold $366$, highest silver $352$), narrow against the $211$-point within-gold spread, so straddling medalists are near-matched on ability---a medal-cutoff natural experiment like that of \citet{agarwalGaule2020}. Bottom-$k$ golds and top-$k$ silvers are statistically indistinguishable ($k{=}10$: $0.078$ vs.\ $0.053$; Mann--Whitney $p \approx 0.2$--$0.5$); a gap appears only once the window widens to re-absorb the recognized tail. The full-tier separation is thus a right-tail phenomenon, not a threshold effect of the medal or the admission it confers: recognition keys on named production, not the institution (the margin test is underpowered, $n \approx 8$--$10$ per side, so a small effect is bounded, not excluded).

The noise floor is calibrated with protocol-matched synthetic-null entities---three fictional NOI~2009 medalists built on the same gold recipe and scored by the same judge---and comes out at essentially zero ($0.000$; three fictional medalists, none recognized by any model): the anti-confabulation provision leaves fictional competitors no recognized facts. The gold-tier mean of $0.104$ is therefore an above-floor signal by an order of magnitude, and the silver ($0.026$) and bronze ($0.019$) means are small but sit clearly above the floor.


\section{Artifact--Creator Pairs: Inversion, Injection, and Asymmetry Detail}
\label{app:asymmetry}

This appendix gives the full tables behind Figure~\ref{fig:inversion} and the Tianshou case study (Section~\ref{sec:results-mechanism}).

\subsection{The inversion and injection tables}

\begin{table}[!htbp]
\centering
\small
\caption{The artifact-over-creator inversion. $\mathrm{NR}_c$, $\mathrm{NR}_a$: creator and artifact recognition; $C\!\to\!A$: fraction of responses about the creator that mention the artifact, $A\!\to\!C$ the reverse. Boldface: artifact exceeds creator. Top block: independent creators; bottom: senior leaders with independent recognition.}
\label{tab:inversion}
\begin{tabular}{lrlrrrr}
\toprule
Creator & $\mathrm{NR}_c$ & Artifact & $\mathrm{NR}_a$ & $\Delta(a{-}c)$ & $C\!\to\!A$ & $A\!\to\!C$ \\
\midrule
Jiayi Weng & 0.22 & \textbf{Tianshou} & \textbf{0.78} & $\mathbf{+0.56}$ & 0.25 & 0.00 \\
Aman Sanger & 0.58 & \textbf{Cursor} & \textbf{0.81} & $\mathbf{+0.23}$ & 1.00 & 0.00 \\
Tri Dao & 0.72 & \textbf{FlashAttention} & \textbf{0.94} & $\mathbf{+0.22}$ & 0.74 & 0.54 \\
Harrison Chase & 0.75 & \textbf{LangChain} & \textbf{0.94} & $\mathbf{+0.19}$ & 1.00 & 0.22 \\
Simon Willison & 0.83 & \textbf{Datasette} & \textbf{1.00} & $\mathbf{+0.17}$ & 0.54 & 0.89 \\
Aravind Srinivas & 0.75 & \textbf{Perplexity} & \textbf{0.89} & $\mathbf{+0.14}$ & 1.00 & 0.33 \\
Dario Amodei & 0.92 & \textbf{Anthropic} & \textbf{1.00} & $\mathbf{+0.08}$ & 1.00 & 0.41 \\
\midrule
Demis Hassabis & 1.00 & Google DeepMind & 1.00 & $0.00$ & 1.00 & 0.62 \\
Andrej Karpathy & 0.86 & nanoGPT & 0.78 & $-0.08$ & 0.05 & 0.76 \\
Mira Murati & 0.78 & Thinking Machines Lab & 0.17 & $-0.61$ & 0.97 & 1.00 \\
\bottomrule
\end{tabular}
\end{table}

\begin{table}[!htbp]
\centering
\small
\caption{Context-injection experiment. $\Delta$ is the creator's recognition change when the artifact name is added to the context (configuration B $-$ A). Under the recognition judge the lift is essentially zero, positive only for the two least-known creators.}
\label{tab:injection}
\begin{tabular}{llr}
\toprule
Creator & Artifact injected & $\Delta$ \\
\midrule
Jiayi Weng & Tianshou & $+0.11$ \\
Simon Willison & Datasette & $+0.08$ \\
Tri Dao & FlashAttention & $+0.06$ \\
Aman Sanger & Cursor & $+0.03$ \\
Harrison Chase & LangChain & $0.00$ \\
Demis Hassabis & Google DeepMind & $0.00$ \\
Andrej Karpathy & nanoGPT & $-0.03$ \\
Mira Murati & Thinking Machines Lab & $-0.08$ \\
Aravind Srinivas & Perplexity & $-0.08$ \\
Dario Amodei & Anthropic & $-0.11$ \\
Lilian Weng & lilianweng.github.io & $-0.11$ \\
\midrule
Mean over $11$ pairs & & $-0.01$ \\
\bottomrule
\end{tabular}
\end{table}

\textbf{The inversion is observational, not inducible.} Under a graded coverage score, an earlier version of this experiment showed an apparent positive lift---but that lift was the model echoing the injected artifact name back into a response that then earned coverage credit, not new stored knowledge. The recognition judge closes exactly that channel: the injected clause is part of the context, so a response that merely restates it earns nothing, and only facts the model knew independently count. Scored this way, the mean lift is essentially zero ($-0.01$), positive only for the two least-known creators (Jiayi Weng, Tri Dao) and negative or null everywhere else. Naming the artifact does not manufacture recognition of the creator. The artifact-over-creator inversion is therefore a property of how attribution is distributed in the corpus---a probe about a famous artifact often omits its creator---not an effect a prompt intervention can induce, which is the stronger and cleaner reading of the mechanism.

\subsection{The Tianshou case}

\begin{table}[!htbp]
\centering
\small
\caption{The Tianshou case: Jiayi Weng against OpenAI individual-contributor peers who share the affiliation but not a widely-named artifact. Tianshou out-ranks him; peers without such an artifact cluster near the floor. (Trevor Blackwell, a YC co-founder, carries independent recognition.)}
\label{tab:tianshou-case}
\begin{tabular}{lc}
\toprule
Entity & Recognition \\
\midrule
\textbf{Jiayi Weng} & \textbf{0.22} \\
\quad Tianshou (his artifact) & 0.78 \\
\midrule
Vicki Cheung & 0.39 \\
Pamela Vagata & 0.33 \\
Alex Wei & 0.11 \\
Andrea Vallone & 0.08 \\
Adam Fry & 0.03 \\
Aaditya Singh & 0.00 \\
\midrule
Peer mean (anonymous ICs) & 0.16 \\
\bottomrule
\end{tabular}
\end{table}

The case distills the mechanism to one person: Jiayi Weng's recognition ($0.22$) sits in the range of his OpenAI IC peers, but his named artifact Tianshou ($0.78$) vastly exceeds him---the recognition attaches to the tool, and reaches the person only through it. The peers who never shipped a widely-named artifact cluster near the floor. Per-model mention data for all $11$ verified pairs (fraction of non-refusal responses that name the counterpart) is released as a supplementary CSV; the models that recognize Jiayi Weng are exactly those whose response names Tianshou.

\textbf{The boundary: reach without name propagation (Section~\ref{sec:results-cases}).} Probed on the same panel, a uniquely-named, GitHub-documented CTF (Hackergame) out-recognizes a course-review site (icourse.club) and the visa-monitoring site tuixue.online, all far below the named-artifact anchor nanoGPT. The ordering tracks whether the site's name travels together with documented, English-readable text, not its user base---the same attribution mechanism, at the boundary where an artifact spreads functionally but anonymously.


\section{Bibliometric Predictors}
\label{app:external}
\label{app:bibliometric}

Full bibliometric-regression results for NameRank on the OpenAlex long-tail researcher cohort ($n = 603$ with complete bibliometrics, citations $500$--$30{,}000$). \texttt{fractional\_citations} divides each paper's citations by its author count; \texttt{first\_last\_citations} counts only papers where the researcher is first or last author; \texttt{n\_works} is the raw count of works; \texttt{mean\_authors} is the mean author count per paper. All $R^2$ on $\log(1+x)$-transformed predictors.

\begin{table}[!htbp]
\centering
\small
\caption{Sole-predictor $R^2$ on recognition. $h$-index, raw citations, and the attribution-weighted variants all cluster around a fifth of the variance; author count carries no signal.}
\label{tab:bibliometric}
\begin{tabular}{lr}
\toprule
Predictor & Sole $R^2$ \\
\midrule
$\log(\text{fractional citations})$ & 0.26 \\
$\log(\text{first/last-author citations})$ & 0.24 \\
$\log(h\text{-index})$ & 0.22 \\
$\log(\text{raw citations})$ & 0.22 \\
$\log(n_\text{works})$ & 0.16 \\
$\log(\text{mean authors/paper})$ & 0.00 \\
\bottomrule
\end{tabular}
\end{table}

The predictors are close substitutes: attribution-weighted fractional citations edge ahead, but $h$-index and raw citations tie near $R^2 \approx 0.22$, and the joint citation-plus-$h$-index model reaches only $0.28$---each adds ${\approx}0.06$ beyond the other because the two are correlated ($r \approx 0.6$). Recognition rises near-monotonically across $h$-index deciles (Table~\ref{tab:h-index-decile}), from $0.18$ at the bottom decile to $0.60$ at the top, but never approaches a ceiling that would let a bibliometric stand in for recognition: every predictor leaves roughly three-quarters of the variance unexplained. That residual is what the named-artifact and corpus-density channels (Sections~\ref{sec:results-mechanism}, \ref{sec:results-geography}) carry, and it is why bibliometrics alone cannot serve as a recognition proxy.

\begin{table}[!htbp]
\centering
\small
\caption{Mean recognition rate by $h$-index decile within OpenAlex long-tail researchers. The ladder rises near-monotonically from the bottom decile to the top; the bottom decile is floored at the working-researcher floor, not at zero.}
\label{tab:h-index-decile}
\begin{tabular}{rrr}
\toprule
Decile & Mean $h$ & Mean recognition \\
\midrule
D1 & 4 & 0.18 \\
D4 & 18 & 0.38 \\
D7 & 31 & 0.46 \\
D10 & 61 & 0.60 \\
\bottomrule
\end{tabular}
\end{table}

\section{Institution and Country Gradient Detail}
\label{app:geography}

Table~\ref{tab:cs-institution} gives the institution-level view; Table~\ref{tab:cs-country} the country-level aggregation behind Figure~\ref{fig:country}. The matched-citation companion design (Figure~\ref{fig:universities}) separates corpus density from productivity.

\begin{table}[!htbp]
\centering
\small
\caption{CS-faculty recognition by institution (institutions with $n \geq 8$ matched faculty). The top-corpus-density institutions lead; the gradient recurs at country level (Figure~\ref{fig:country}) and holds at matched citations (Figure~\ref{fig:universities}).}
\label{tab:cs-institution}
\begin{tabular}{lrrrr}
\toprule
Institution & $n$ & Recognition \\
\midrule
Stanford University & 13 & \textbf{0.78} \\
University of Washington & 21 & 0.74 \\
Carnegie Mellon University & 50 & 0.71 \\
Princeton University & 15 & 0.67 \\
Cornell University & 17 & 0.65 \\
University of Michigan & 20 & 0.62 \\
\bottomrule
\end{tabular}
\end{table}

\textbf{The matched-citation design.} The faculty-roster comparison above cannot separate corpus density from productivity, because it does not condition on output. A companion design does: sampling researchers into a fixed citation window ($500$--$30{,}000$) stratified by institution, so the compared groups have matched bibliometrics by construction. At matched citations the top-corpus-density institutions still lead---MIT and Berkeley faculty out-recognize UCSD and Irvine faculty at the same citation level (Figure~\ref{fig:universities}b)---so the gradient is corpus density, not research output.

\begin{table}[!htbp]
\centering
\small
\caption{CS-faculty recognition by country of institutional affiliation, for countries with $\geq 8$ sampled faculty. The firm contrast is the USA against China (non-overlapping intervals); mid-sized samples (UK, Hong Kong, Canada) should not be finely ranked.}
\label{tab:cs-country}
\begin{tabular}{lrr}
\toprule
Country & $n$ & Recognition \\
\midrule
USA & 206 & \textbf{0.64} \\
UK & 10 & 0.64 \\
Hong Kong & 8 & 0.59 \\
Canada & 8 & 0.58 \\
\textbf{China} & 8 & \textbf{0.33} \\
\bottomrule
\end{tabular}
\end{table}

The well-sampled core of the gradient is the USA (recognition $0.64$) against China ($0.33$), whose $95\%$ confidence intervals do not overlap. The mechanism is English-corpus density: identical research output propagates through a denser English-language ecosystem for a US-based researcher than for a peer whose output travels mainly through the papers themselves.


\section{Cross-Language Robustness}
\label{app:cross-lang}

\textbf{Sub-run cohort selection.} The Chinese-prompt sub-run covers $240$ entities: the full diagnostic reference set ($37$), all NOI China gold medalists ($29$), all DeepSeek-V3 paper authors ($69$), $30$ random samples each from the CMO, CPhO, and MSRA Fellowship cohorts, and $5$ each from the writer, filmmaker, and politician cohorts as Western controls---$240 \times 36$ records.

\begin{table}[!htbp]
\centering
\small
\caption{Cross-language NameRank by cohort: $\Delta = \mathrm{NR}_\mathrm{zh} - \mathrm{NR}_\mathrm{en}$ on the $240$-entity Chinese-prompt sub-run. Every cohort delta is within $\pm 0.04$.}
\label{tab:cross-lang}
\begin{tabular}{lrrrr}
\toprule
Cohort & $n$ & $\mathrm{NR}_\mathrm{en}$ & $\mathrm{NR}_\mathrm{zh}$ & $\Delta$ \\
\midrule
MSRA PhD Fellowship & 30 & 0.166 & 0.183 & $+0.018$ \\
DeepSeek-V3 authors & 69 & 0.069 & 0.075 & $+0.006$ \\
CMO China gold & 30 & 0.029 & 0.028 & $-0.001$ \\
NOI China gold & 29 & 0.086 & 0.084 & $-0.002$ \\
CPhO China first prize & 30 & 0.034 & 0.029 & $-0.006$ \\
Diagnostic reference set & 37 & 0.627 & 0.615 & $-0.012$ \\
Writers (control) & 5 & 0.906 & 0.872 & $-0.033$ \\
Filmmakers (control) & 5 & 0.994 & 0.961 & $-0.033$ \\
Politicians (control) & 5 & 0.900 & 0.861 & $-0.039$ \\
\bottomrule
\end{tabular}
\end{table}

\textbf{The prompt-language effect is essentially null.} This is itself a finding, and a reassuring one for a cross-lingual instrument: switching the probe from English to Chinese changes no cohort's recognition rate by more than $0.04$, and the Chinese-named credential cohorts (DeepSeek authors $+0.006$, NOI $-0.002$, CMO $-0.001$) that a language-matching artifact would most inflate move least. An earlier graded-coverage version of this sub-run showed a sizeable Chinese-prompt lift for Chinese-named cohorts; that lift was language-matched \emph{verbosity}---a Chinese prompt elicits a longer Chinese answer that scores more coverage points---not extra knowledge. Once the judge credits only a verified non-guessable fact, the wording of the question stops mattering, because whether the model knows Lin Yankai is a property of its weights, not of the language it is asked in.

\textbf{Per-model prompt-language deltas.} The residual shift is small and not vendor-specific: the mean absolute per-model delta is $0.023$, and the two model classes are indistinguishable in aggregate (Western $-0.005$, Chinese $+0.002$). The largest per-model movers---deepseek-v4-pro-think $+0.08$, gemma-3-12b $+0.08$; llama-3.1-8b $-0.05$, gemma-3-4b $-0.04$---are small-model noise on either side, not a systematic home-language advantage.

\textbf{Per-entity deltas.} The full per-entity $\Delta(\mathrm{NR}_\mathrm{zh} - \mathrm{NR}_\mathrm{en})$ table is released as a supplementary CSV. Even the extremes are modest---the largest suppressions are Harrison Chase $-0.19$, Aleksander Madry $-0.14$, and Canyi Lu $-0.14$, all English-context entities where a few borderline models drop the fact on the Chinese prompt:

\begin{center}
\small
\begin{tabular}{lrrrr}
\toprule
Entity & $\mathrm{NR}_\mathrm{en}$ & $\mathrm{NR}_\mathrm{zh}$ & $\Delta$ & Cohort \\
\midrule
Jiayi Weng & 0.22 & 0.22 & $+0.00$ & reference set \\
Tri Dao & 0.72 & 0.72 & $+0.00$ & reference set \\
Andrej Karpathy & 0.86 & 0.89 & $+0.03$ & reference set \\
FlashAttention & 0.86 & 0.92 & $+0.06$ & reference set \\
Tianshou & 0.83 & 0.89 & $+0.06$ & reference set \\
Bojie Li & 0.22 & 0.19 & $-0.03$ & reference set \\
\midrule
Pan Zhou & 0.08 & 0.22 & $+0.14$ & MSRA Fellowship \\
Lin Yankai & 0.06 & 0.17 & $+0.11$ & NOI China gold \\
Chen Jianfei & 0.03 & 0.14 & $+0.11$ & NOI China gold \\
Yingwei Pan & 0.17 & 0.28 & $+0.11$ & MSRA Fellowship \\
\bottomrule
\end{tabular}
\end{center}

The residual pattern, where it exists at all, is confined to \emph{low-recognition} Chinese-named credential holders (bottom block): for a name the panel barely knows, a Chinese prompt occasionally tips a borderline model over the recognition threshold, worth at most ${\sim}0.14$. For every entity the panel actually recognizes---the reference set---the delta is within $\pm 0.06$. Name attribution that has genuinely propagated is stored language-independently; the prompt language cannot manufacture it and barely nudges it.


\section{News-Event Cohort: Construction and Detail}
\label{app:events}

This appendix documents the construction of the $258$-event cohort of Section~\ref{sec:results-events} and reports the full tables behind its findings. All construction scripts, inputs, and derived data are released under \texttt{experiments/t4\_1\_news\_events/} in the repository.

\subsection{Sampling frame and filters}

Candidate events are harvested from the ``Events'' sections of the English Wikipedia year articles (\emph{2021}, \emph{2022}, \emph{2023}) and of $32$ country-year articles (``\emph{2021 in India}'', ``\emph{2022 in Nigeria}'', \ldots) chosen to balance nine world regions ($2{,}245$ unique linked articles). Each candidate is classified by Gemini~3 Flash Preview (the study's judge family; temperature $0$, structured output) from its title, short description, and first paragraph into: discrete event or not, start year and month, primary country, region, and category. An article qualifies as a \emph{discrete event} if it describes a specific occurrence or tightly bounded series (an election, a storm, a trial, a tournament, an accident, a protest wave) that began at an identifiable date; people, organizations, products, laws, ongoing multi-year topics, and list/timeline articles are excluded ($1{,}041$ qualify). Post-classification filters require a standard-type article with an intro of $\geq 40$ words and a first-year pageview record of $\geq 1{,}000$ total views over $\geq 200$ recorded days; $655$ events survive. Stratifying by $\log_{10}$(total views) band ($<10^4$ through $\geq 10^7$) crossed with eight region groups, capped at $14$ events per cell, yields the released cohort of $258$.

\subsection{Attention measures and cross-checks}

For each event we retrieve the daily en-Wikipedia pageview series of its article (Wikimedia REST API, user traffic only) for the $365$ days from the first day of the event's start month, and define \textbf{total attention} (summed daily views), \textbf{peak attention} (maximum daily views), and \textbf{effective duration} $=$ total$/$peak in days, linked by the identity $\log(\text{total}) = \log(\text{peak}) + \log(\text{duration})$. Wikipedia pageviews are a standard collective-attention proxy~\citep{mestyan2013early,ripberger2011capturing}; we prefer them to Google Trends because the pageview record is absolute, archived, and reproducible, whereas the public Trends interface returns only normalized indices under heavy rate limits. Two caveats. First, the ledger is attention \emph{to the event's English Wikipedia article}, so all matched-attention comparisons condition on English-audience interest; this is a feature for the region analysis (it holds English exposure fixed) but means globally quiet, locally enormous events enter as quiet. Second, attention can fragment across sibling articles (\emph{Killing of Tyre Nichols} vs.\ \emph{Tyre Nichols protests}), adding measurement noise that biases dose--response estimates toward zero. A GDELT news-coverage-volume cross-check~\citep{leetaru2013gdelt} on the event's name phrase over the same window is reported under Full Results below and released with the data.

\subsection{Probes and panel}

Golds are the article's intro truncated to $100$--$200$ words at a sentence boundary; disambiguating contexts are generated deterministically from the classification fields only---category phrase, location, month, year---never from article content, mirroring the role/affiliation/field rule of the person cohorts. Probe template, judge, refusal detection, and scoring are the same $36$-model recognition pipeline as every other cohort (Section~\ref{sec:method}).

\textbf{Panel invariance.} The event window ($2021$--$2023$) sits inside every panel model's training cutoff, so recognition failures cannot be corpus-timing artifacts. Because the entity factor dominates the verdict variance (Appendix~\ref{app:scoring}), the event ordering is a property of the events, not the panel: splitting the panel by vendor family or by release vintage reproduces the same ranking, the disjoint-panel robustness check applied to the one cohort whose exposure side is externally recorded.

\textbf{Cutoff coverage.} Three panel models have training cutoffs inside the final months of the event window (Mistral-Small-24b, $2023$-$10$; Llama-3.1-8b and Llama-3.3-70b, $2023$-$12$); the eight replacement models are $2026$ releases with cutoffs far past the window. Neither exclusion changes the peak-dominant decomposition: dropping the $22$ events that begin in October--December $2023$, and separately dropping the three near-cutoff models for all events, both leave peak the dominant standardized predictor (${\approx}+0.12$--$0.13$) and duration marginal (${\approx}+0.04$), matching the headline in Table~\ref{tab:events-deciles}'s regression.

\subsection{Full results}

\textbf{Dose--response.} Table~\ref{tab:events-deciles} gives the decile ladder: recognition rises monotonically from $0.61$ in the lowest pageview decile to $0.95$ in the highest, and the refusal rate falls in lockstep from $0.12$ to $0.01$. Sole-predictor $R^2$ on recognition: $\log$ total views $0.387$ (repeated $10$-fold CV $0.37 \pm 0.16$), $\log$ peak $0.342$ ($0.33 \pm 0.16$), $\log$ tail interest (mean daily views in days $300$--$365$) $0.345$, and $\log$ effective duration essentially zero ($0.001$). Attention explains recognition markedly better than it explained the earlier graded coverage score---a recorded attention ledger is a strong external predictor of whether a name propagates. Gold length is a partial confound (it alone explains $0.24$, since bigger events carry longer intros), but attention dominates it: adding gold length to the attention model lifts $R^2$ only from $0.387$ to $0.43$, and the attention slope survives at $+0.10$ per decade (se $0.013$).

\begin{table}[!htbp]
\centering
\small
\caption{Recognition by decile of total first-year pageviews ($n = 258$). Attention converts refusal into recognition: the recognition rate climbs and the refusal rate falls monotonically across the range.}
\label{tab:events-deciles}
\begin{tabular}{rrrrr}
\toprule
Decile & $n$ & Geo.\ mean views & Recognition & Refusal \\
\midrule
D1 & 25 & 2{,}415 & 0.61 & 0.12 \\
D2 & 26 & 7{,}395 & 0.62 & 0.11 \\
D3 & 26 & 16{,}514 & 0.66 & 0.12 \\
D4 & 26 & 39{,}348 & 0.70 & 0.07 \\
D5 & 26 & 67{,}583 & 0.77 & 0.06 \\
D6 & 25 & 133{,}127 & 0.80 & 0.06 \\
D7 & 26 & 251{,}374 & 0.83 & 0.04 \\
D8 & 26 & 480{,}323 & 0.92 & 0.01 \\
D9 & 26 & 1{,}223{,}071 & 0.93 & 0.01 \\
D10 & 26 & 3{,}312{,}033 & 0.95 & 0.01 \\
\bottomrule
\end{tabular}
\end{table}

\textbf{Peak vs.\ duration.} Joint regression of recognition on the standardized log-components: peak $+0.127$ (HC1 se $0.011$), duration $+0.043$ (se $0.012$), joint $R^2 = 0.387$ (CV $0.37 \pm 0.17$); the marginal $R^2$ of duration given peak is only $0.045$, whereas peak alone explains $0.34$. So recognition loads on how \emph{intense} the attention was, not how long it lasted: a name is minted into the corpus by a high-water-mark of coverage, and stretching that coverage over more days adds little. Within peak terciles, moving from the shortest to the longest duration tercile moves recognition by $+0.01$, $+0.14$, and $+0.03$ across the three terciles---a weak, non-monotone duration effect at matched peak, concentrated in the mid-peak band.

\textbf{Recurring-series flag.} An event is a series instance if substituting a nearby year (or adjacent ordinal/Roman edition marker) into its title yields another standard Wikipedia article; the detector sweeps $y \pm 1..6$ to cover $4$--$6$-year election cycles. $106$ of $258$ events flag as recurring. Under the recognition verdict the recurring penalty is \emph{null}: the raw recurring-vs-singular gap ($0.83$ vs.\ $0.75$) reverses sign once attention is controlled, leaving an adjusted delta of $+0.006$ (se $0.019$). Whatever a recurring title costs in graded coverage, it does not change whether the model recognizes the specific edition.

\textbf{Region and vintage.} Table~\ref{tab:events-region} reports raw and attention-adjusted region contrasts. Event-year effects at matched attention show a clear recency penalty: $2022$ vs.\ $2021$ is $-0.033$ (se $0.020$), and $2023$ vs.\ $2021$ is $-0.132$ (se $0.022$)---the most recent events are the least recognized at matched attention, consistent with post-event derivative text still accumulating and with several panel cutoffs falling inside 2023.

\textbf{GDELT cross-check.} Of the $258$ name-phrase queries, the rate-limited GDELT DOC API returned data for $159$; $118$ had nonzero matched coverage. On those, $\log$ GDELT coverage volume correlates with $\log$ pageviews at $r = 0.33$---the two ledgers partially agree---and predicts recognition at $R^2 = 0.09$, weaker than the pageview ledger's $0.34$--$0.39$ but non-trivial. The gap is mostly query noise: GDELT matches the event's literal name phrase, which misses paraphrase references and truncates awkwardly for long titles, and the returned subset is rate-limit-selected rather than random. We therefore rely on the pageview ledger and report GDELT as a consistency check on the attention record, not an independent predictor.

\begin{table}[!htbp]
\centering
\small
\caption{Region groups: raw mean NameRank and the adjusted delta vs.\ the Anglo-America \& Oceania baseline after controlling $\log_{10}$ total views (HC1 standard errors). The single Global-classified event (the Queen's Platinum Jubilee Beacons) is omitted.}
\label{tab:events-region}
\begin{tabular}{lrrr}
\toprule
Region group & $n$ & Raw mean & Adj.\ $\Delta$ (se) \\
\midrule
Anglo-America \& Oceania & 45 & 0.786 & (baseline) \\
East Asia & 31 & 0.797 & $+0.025$ (0.037) \\
Eastern Europe \& Russia & 28 & 0.823 & $+0.020$ (0.031) \\
Middle East \& Africa & 35 & 0.824 & $+0.014$ (0.035) \\
Latin America & 32 & 0.763 & $-0.012$ (0.031) \\
Western Europe & 44 & 0.780 & $-0.015$ (0.033) \\
South \& Southeast Asia & 42 & 0.697 & $-0.068$ (0.032) \\
\bottomrule
\end{tabular}
\end{table}


\section{Self-Reported Recognition: Design and Detail}
\label{app:selfreport}

This appendix gives the design, estimation, and full results behind Section~\ref{sec:results-selfreport} (released as \texttt{experiments/t5\_4\_self\_report/}).

\subsection{Design}

\textbf{Entities.} $430$ total: $400$ real entities sampled from the full study, stratified over six panel-NameRank bands ($[0,.05)$, $[.05,.15)$, $[.15,.30)$, $[.30,.50)$, $[.50,.70)$, $[.70,1]$ with quotas $40/40/70/90/90/70$ and at most $12$ per cohort), plus the $30$ fictional entities of Appendix~\ref{app:synthetic} as traps.

\textbf{Pairs.} $3{,}350$ unique pairs in five strata: $2{,}200$ with both entities in the recognized region (panel NameRank $\ge 0.15$; half constrained to close gaps $|\Delta| < 0.15$), $500$ crossing the recognition boundary ($\ge 0.30$ vs.\ $< 0.05$), $250$ with both below $0.15$, and $400$ trap pairs (each fictional entity against $12$ real partners spread over the bands, plus $40$ fictional--fictional pairs). A further $335$ pairs ($10\%$) are re-presented in swapped order as a position-bias probe, giving $3{,}685$ presentations per model. Presentation order (A/B) is randomized once and shared across models.

\textbf{Probe.} Each pair is shown with the study's disambiguation contexts, and the model is asked which entity it knows more about---``specific, verifiable facts such as works, roles, achievements, or dates''---answering exactly one of \texttt{A}, \texttt{B}, \texttt{EQUAL} (knows both to similar depth), or \texttt{NEITHER} (recognizes neither). The three self-report models are the panel checkpoints GPT-5.5, Claude Opus 4.6, and Gemini 3.1 Pro at standard settings, so each model's verdicts can be compared against its \emph{own} per-record recognition behavior. All $11{,}055$ calls returned a parseable verdict.

\textbf{Estimation.} Each model's \texttt{A}/\texttt{B}/\texttt{EQUAL} verdicts on real--real pairs are fit with Davidson's tie-extended Bradley--Terry model \citep{bradley1952rank,davidson1970ties}---the batch analogue of the Chatbot Arena aggregation \citep{chiang2024chatbot}---by ridge-regularized maximum likelihood; \texttt{NEITHER} verdicts are treated as abstentions. Confidence intervals are $95\%$ normal-approximation pair-bootstrap ($200$ refits). Entity-level statistics use entities with at least five decided comparisons. Per-model labels for the binary analyses come from the model's own recognition record: \emph{known} $=$ score $\ge 0.15$ and no refusal; \emph{unknown} $=$ refusal or score $< 0.05$.

\subsection{Results}

\begin{table}[!htbp]
\centering
\small
\caption{Self-reported recognition per model. $\rho_{\text{panel}}$: Spearman correlation of the Bradley--Terry self-ranking with panel NameRank over real entities (bootstrap $95\%$ CI). Dir.\ acc.: fraction of decided verdicts agreeing with the panel ordering on both-known pairs with $\Delta\text{NameRank} \ge 0.15$. Boundary: known-vs-unknown pairs by the model's own labels---share picking the known side / the unknown side / abstaining. Trap: share of fictional-vs-known pairs where the fictional side was preferred (ties included). Reversal: verdict consistency under A/B swap.}
\label{tab:selfreport}
\begin{tabular}{lccccc}
\toprule
Model & $\rho_{\text{panel}}$ & Dir.\ acc. & Boundary (pick/err/abstain) & Trap & Reversal \\
\midrule
GPT-5.5          & $0.61$ {\scriptsize$[.59,.63]$} & $0.82$ & $0.74$ / $0.12$ / $0.14$ & $1/232$ & $0.88$ \\
Claude Opus 4.6  & $0.53$ {\scriptsize$[.50,.56]$} & $0.81$ & $0.63$ / $0.14$ / $0.23$ & $0/241$ & $0.85$ \\
Gemini 3.1 Pro   & $0.57$ {\scriptsize$[.55,.59]$} & $0.81$ & $0.72$ / $0.09$ / $0.16$ & $0/211$ & $0.90$ \\
\bottomrule
\end{tabular}
\end{table}

\textbf{Partial reconstruction of the panel ordering.} Table~\ref{tab:selfreport}: the self-rankings correlate $\rho \approx 0.53$--$0.61$ per model with panel recognition (the aggregate self-report scale, averaging the three, reaches $\rho \approx 0.83$) and agree with the panel ordering on $81\%$ of decided both-known pairs at gaps $\ge 0.15$. Position bias is negligible (reversal consistency $0.85$--$0.90$; first-position win rate $0.46$--$0.49$).

\textbf{What self-report reads.} Panel NameRank factors into a \emph{prevalence} component (the share of the $36$ models that know the entity, i.e.\ score $\ge 0.15$) and a \emph{conditional level} (mean score among the models that know it). The self-rankings load on prevalence ($\rho = 0.72$--$0.76$) and only weakly on conditional level ($\rho = 0.11$--$0.19$); the models' own behavioral scores do the reverse (conditional level $\rho = 0.60$--$0.71$, prevalence $\rho = 0.07$--$0.15$ among known entities). Consequently the self-ranking is nearly uncorrelated with the model's own scores among entities it knows ($\rho = +0.04$ / $-0.06$ / $-0.09$), while the three vendors' self-rankings agree pairwise at $\rho = 0.90$--$0.91$---each model agrees with the \emph{other vendors' introspection} far better than with its \emph{own behavior}. The conditional level is gold-conditional rather than fame-driven (specific dense golds are harder to cover than boilerplate golds; cf.\ the level-sensitivity in Appendices~\ref{app:prompt-sensitivity} and~\ref{app:synthetic}), which is also why directional accuracy against raw own-score gaps lands below chance ($0.34$--$0.40$): raw per-model scores are not a cross-entity depth scale, and we do not use them as one.

\textbf{Fame-incongruent boundary pairs.} Restricting known-vs-unknown pairs to those where corpus fame contradicts the model's own state (the unknown entity has the higher panel NameRank by $\ge 0.05$): abstention rises from $10$--$20\%$ (congruent) to $35$--$64\%$, the famous-but-unknown side is claimed in only $0$--$9\%$ of verdicts, and only GPT-5.5 still sides with its own knowledge on most such pairs ($0.65$, vs.\ $0.36$ for both others; $n = 26/14/69$, so these splits are indicative). Self-report degrades into abstention, not fame-following fabrication.

\textbf{Trap arm.} Across $684$ fictional-vs-known pairs the fictional entity was preferred once (GPT-5.5, $0.4\%$ of its $232$; ties included); \texttt{NEITHER} was answered on $95$--$100\%$ of fictional--fictional pairs. Recognition claims are near-perfectly precise; the error budget is spent on under-claiming (abstention on $14$--$23\%$ of genuine boundary pairs).

\textbf{Caveats.} The behavioral ground truth is one judged response per (entity, model), so response-level noise attenuates all within-model comparisons; the prevalence/conditional decomposition and the panel-side statistics are unaffected. \texttt{EQUAL} usage varies by vendor ($3\%$ of GPT-5.5 verdicts, $5\%$ Claude, $12\%$ Gemini), which the tie-extended likelihood absorbs. The experiment covers three frontier models; smaller models' self-reports may behave differently.


\section{Scoring-Rule Validity: Variance, Refusals, and Embeddings}
\label{app:scoring}
\label{app:variance}

This appendix and Appendices~\ref{app:prompt-sensitivity}--\ref{app:confounds} give the evidence behind the validation battery of Section~\ref{sec:results-validation}, summarized threat-by-threat in Table~\ref{tab:threats}. The remainder of this appendix collects the evidence that the scoring rule---an LLM recognition judge over a panel---measures the entity rather than the panel, and that no simpler rule would do.

\begin{table}[htbp]
\centering
\small
\caption{The validation battery: each threat to the findings, the check run against it, and the outcome. Full tables and figures in the referenced appendices.}
\label{tab:threats}
\begin{tabular}{>{\raggedright\arraybackslash}p{3.4cm}>{\raggedright\arraybackslash}p{3.6cm}>{\raggedright\arraybackslash}p{5.6cm}c}
\toprule
\textbf{Threat} & \textbf{Check} & \textbf{Outcome} & \\
\midrule
Score reflects the panel, not the entity & Variance decomposition; disjoint-panel re-score of the events & Entity variance ${\approx}6\times$ model variance; a replacement panel reproduces the event ordering & App.~\ref{app:scoring} \\
Fluent hallucinations inflate scores & Recognition verdict on the full panel & A confident wrong-person biography scores zero; the never-refusing hallucinator cluster earns nothing & App.~\ref{app:scoring} \\
An embedding score would do instead & Judge vs.\ embedding comparison & Embedding similarity is flat across recognition and credits hallucinations & App.~\ref{app:scoring} \\
Just ask the model (self-report) & Pairwise self-report on three frontier models & Self-report reads the shared corpus prior, not the model's own state---a prior, not a measurement & App.~\ref{app:selfreport} \\
An artifact of probe wording & Four-template paraphrase re-probe & Wording explains ${\approx}0\%$ of variance; ladder preserved ($\rho \geq 0.997$), only levels shift & App.~\ref{app:prompt-sensitivity} \\
An artifact of probe language & Chinese re-probe of $240$ entities & No cohort moves by more than $0.04$; recognition is stored language-independently & App.~\ref{app:cross-lang} \\
The silent zone is corpus timing, not obscurity & Training-cutoff natural experiment & Matched DiD negative---recent cohorts gain \emph{less} than capability drift; silence is intrinsic & App.~\ref{app:cutoff} \\
Findings depend on the judge's family & Three-family re-judge & Ranking identical across judges; GPT-5.1 stricter, but the uniform offset moves no relative finding & App.~\ref{app:cross-judge} \\
Unknown names score above zero & Synthetic-null entities through the full pipeline & Floor ${\approx}0.01$ (people/artifacts), ${\approx}0.06$ (paper titles); floor-adjusting keeps the credential gap & App.~\ref{app:synthetic} \\
Gold, context leakage, Wikipedia, or web attention drive the results & Ablations; long-window pageview baseline & All rejected; not a Wikipedia flag nor a pageview rank & App.~\ref{app:confounds} \\
\bottomrule
\end{tabular}
\end{table}

\subsection{The score measures the entity, not the panel}

Decomposing the variance of the per-record recognition verdicts (Table~\ref{tab:variance}), the entity factor explains roughly six times the variance the model factor does. Model identity contributes a small share, and the per-model generosity differences that produce it (frontier reasoning models at the generous end, small hallucination-averse models at the strict end) are tabulated in full in the \repodoc{per-model-stats.md}{released per-model statistics}. This is the quantitative form of the disjoint-panel robustness check: an entirely separate replacement panel reproduces the event-cohort ordering, because the score is a property of the entity, not the fleet.

\begin{table}[!htbp]
\centering
\small
\caption{Variance decomposition of the per-record recognition verdicts ($36$-model panel). Entity and cohort are nested, so each row is the share of total variance the factor independently explains. The entity dominates.}
\label{tab:variance}
\begin{tabular}{lr}
\toprule
Source & \% of verdict variance \\
\midrule
\textbf{Entity} & \textbf{56\%} \\
Cohort & 37\% \\
Model & 9\% \\
\bottomrule
\end{tabular}
\end{table}

\subsection{Refusal structure and why a recognition verdict is required}

The refusal rate---the fraction of the panel answering ``unknown''---inverts recognition at cohort level: silent cohorts draw high refusal rates, universal cohorts near zero. Models honestly refuse on low-recognition cohorts rather than hallucinating, the threshold-cleanliness property surfacing in raw response behavior.

The per-model refusal signature shows why a recognition verdict, not a coverage score, is essential. The panel splits into a high-refusal cluster (hallucination-averse models) and a zero-refusal cluster of \emph{fluent hallucinators} that always produce a confident-looking biography (\repodoc{per-model-stats.md}{released per-model statistics}). The recognition verdict handles the latter directly: a confident wrong-person biography states no fact that is true of the target entity, so it scores non-recognition regardless of how fluent or topically on-target it is. A coverage-based or refusal-based metric would systematically over-credit the fluent-hallucinator cluster; the verdict does not.

\subsection{Embedding similarity cannot replace the judge}

Embedding cosine similarity between response and gold shifts only modestly across the recognition boundary (Table~\ref{tab:embedding}): recognized responses embed a little closer to the gold ($0.88$ vs.\ $0.80$), but the distributions overlap so heavily that no threshold separates recognition from fluent hallucination. Embeddings measure topical similarity, which is high even for hallucinated bios---a fabricated biography of a CS researcher still embeds close to a gold about that researcher's actual subfield. The most extreme disagreements are confident, entirely fabricated bios of common-name long-tail researchers, exactly the fluent-hallucination case the judge exists to catch. An embedding-only NameRank would credit these as recognized.

\begin{table}[!htbp]
\centering
\small
\caption{Mean gold-response embedding similarity for recognized vs.\ non-recognized non-refusal responses. The two overlap heavily ($0.88$ vs.\ $0.80$), so a similarity threshold cannot substitute for the judge.}
\label{tab:embedding}
\begin{tabular}{lrr}
\toprule
Verdict & $n$ & Mean embedding sim. \\
\midrule
Recognized & 84{,}518 & 0.88 \\
Not recognized (non-refusal) & 43{,}831 & 0.80 \\
\bottomrule
\end{tabular}
\end{table}

\subsection{Per-cohort within-entity disagreement}

The within-entity model disagreement---the spread of recognition verdicts across the $36$-model panel, averaged within cohort---is itself a probe of corpus-pocket distribution. Competition-credential and long-tail academic cohorts show the highest disagreement: different models, trained on different fractions of the available corpus, see different fractions of the entity, so ``which model do you ask?'' changes the answer. Universal-presence cohorts (named methods, conferences, books) show near-total agreement, because the name appears in nearly every training source. The full ladder is in the \repodoc{per-model-stats.md}{released per-model statistics}.


\section{Prompt-Paraphrase Sensitivity}
\label{app:prompt-sensitivity}

The headline metric uses a single fixed probe template (Section~\ref{sec:construct}). The recognition verdict is wording-robust \emph{by construction}, and this is worth making precise because it is the property a graded coverage score lacks. A graded score rewards how much of the gold a response reproduces, so a phrasing that elicits a longer or more fluent answer scores higher even when the model knows nothing more; wording therefore leaks directly into the score. The recognition verdict asks a different question---does the response contain at least one specific, non-guessable fact that is true of the entity and absent from the context?---whose answer does not depend on how the question was phrased or how long the answer runs. Whether the model \emph{knows} such a fact is a property of its weights; a paraphrase of the probe changes what the model \emph{says} but not whether a recognizable fact is available to say. The verdict is thus insulated from the length-and-fluency channel that made graded coverage wording-sensitive.

This is a design guarantee rather than an empirical fit, so it holds for every entity, not on average. The one residual channel a paraphrase can open is the refusal boundary---a differently worded ``say unknown if unsure'' instruction can nudge a borderline model to withhold rather than answer---which affects only entities already sitting at the recognized/not-recognized threshold and cannot move a clearly-recognized or clearly-silent entity.

\textbf{Measured confirmation.} We re-probed a cohort-stratified subset ($136$ entities across $12$ cohorts) on the full $36$-model panel under three paraphrased templates (T1--T3) alongside the original (T0), judged by the same recognition rubric. The four templates:

\begin{verbatim}
T0: Tell me what you know about {name}, who is {context}. If you do
    not recognize this entity, respond with "unknown". Limit your
    response to about 150 words.
T1: What can you tell me about {name}? They are {context}. Reply with
    only "unknown" if you are not familiar with this entity. Aim for
    about 150 words.
T2: Please share what you know about {name}, {context}. If unknown to
    you, answer with just "unknown". Keep it under ~150 words.
T3: Briefly describe {name}, who is {context}. Respond "unknown" if
    you have no information. Around 150 words please.
\end{verbatim}

The prediction holds sharply. In an additive variance decomposition of the per-record verdict over the balanced ($4{,}680$-cell) design, \textbf{template explains essentially none of the variance} (Table~\ref{tab:prompt-variance}), against $53\%$ for the entity and $11\%$ for the model. The four templates assign nearly the same overall recognition rate ($0.50$--$0.53$), and the per-cohort ladder is reproduced under every wording (Table~\ref{tab:prompt-cohort}): the cohort means correlate at $\rho = 0.997$--$0.999$ across template pairs, and the silent$\to$universal ordering is preserved (Spearman $0.99$). Per (entity, model), two templates return the same verdict on $90$--$91\%$ of records; the residual $9$--$10\%$ is concentrated exactly where the design argument places it---at the recognized/not-recognized boundary, on entities near the threshold. Cross-study comparisons should still hold the probe fixed, since that boundary channel shifts absolute levels by a point or two, but no finding depends on the wording.

\begin{table}[!htbp]
\centering
\small
\begin{minipage}[t]{0.42\textwidth}
\centering
\caption{Additive variance decomposition over the paraphrase records. Template is a negligible component beside entity, cohort, and model.}
\label{tab:prompt-variance}
\begin{tabular}{lr}
\toprule
Factor & \% of variance \\
\midrule
Entity & 53\% \\
Cohort & 37\% \\
Model & 11\% \\
\textbf{Template} & \textbf{0\%} \\
\bottomrule
\end{tabular}
\end{minipage}\hfill
\begin{minipage}[t]{0.54\textwidth}
\centering
\caption{Cohort recognition under each template. The ladder is preserved under all four wordings; absolute levels move by at most a point or two.}
\label{tab:prompt-cohort}
\begin{tabular}{lrrrr}
\toprule
Cohort & T0 & T1 & T2 & T3 \\
\midrule
Named methods & 0.95 & 0.93 & 0.96 & 0.95 \\
OSS projects & 0.87 & 0.85 & 0.87 & 0.85 \\
Foundation models & 0.83 & 0.82 & 0.82 & 0.85 \\
Writers & 0.81 & 0.76 & 0.79 & 0.81 \\
CS faculty & 0.56 & 0.52 & 0.54 & 0.54 \\
OpenAlex researchers & 0.37 & 0.34 & 0.37 & 0.37 \\
IMO gold & 0.12 & 0.15 & 0.18 & 0.15 \\
GPT-5 authors & 0.09 & 0.11 & 0.07 & 0.11 \\
\bottomrule
\end{tabular}
\end{minipage}
\end{table}

\section{Training-Cutoff Gradient: The Silent Zone Is Intrinsic}
\label{app:cutoff}

A natural worry about the silent zone (Section~\ref{sec:results-headline}) is that it reflects \emph{corpus timing} rather than \emph{intrinsic obscurity}: perhaps low-NameRank entities are merely too recent for current models to have ingested, and a future fleet would lift them. Two contemporaneously-published cohorts test this directly. The DeepSeek-V3 author cohort became nameable in any indexable corpus only when the technical report was posted (December 2024); the GPT-5 system-card author cohort only at the system card's publication (August 2025). The panel's training cutoffs straddle both dates (released with the data; cutoffs span 2023-10 to 2026-03), so a model whose cutoff predates an emergence date \emph{cannot} have ingested the document.

\begin{figure}[!htbp]
    \centering
    \includegraphics[width=\textwidth]{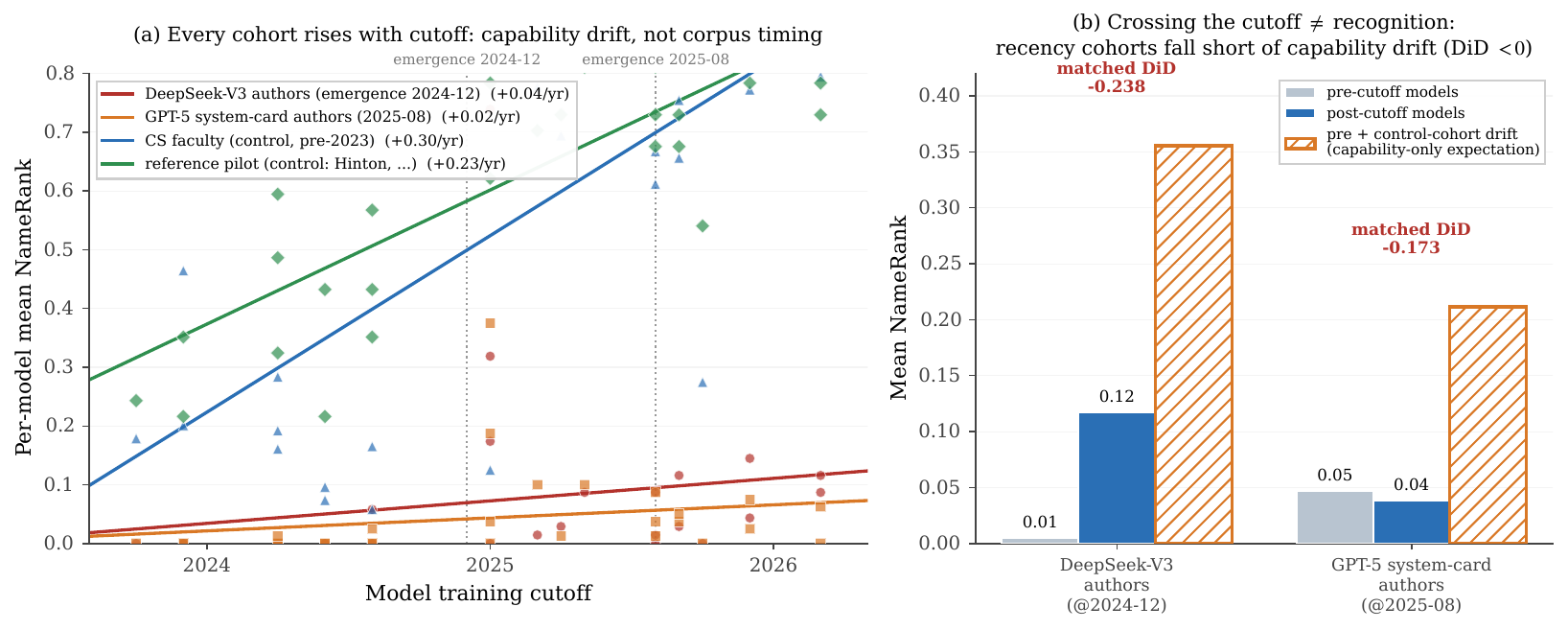}
    \caption{Training-cutoff gradient. \textbf{(a)}~Per-model mean NameRank vs.\ training cutoff for two ``recency'' cohorts (DeepSeek-V3 authors, GPT-5 system-card authors) and two stable controls; every cohort rises at ${\sim}0.1$--$0.3$/yr, a model-capability confound. \textbf{(b)}~The recency cohorts' pre/post-emergence jump is smaller than the same split on the stable controls, so the matched difference-in-differences is negative ($-0.24$, $-0.17$).}
    \label{fig:cutoff-gradient}
\end{figure}

The result (Figure~\ref{fig:cutoff-gradient}) separates two effects. First, a model-vintage confound dominates the cutoff axis: \emph{every} cohort's recognition rises with training cutoff at roughly $0.1$--$0.3$ per year, including cohorts (containing, e.g., Geoffrey Hinton) nameable long before any cutoff---capability and willingness-to-commit drift, not corpus growth. Second, once that drift is netted out by matched difference-in-differences against stable control cohorts, no corpus-timing boost remains---if anything the recency cohorts \emph{lag} the drift: the DeepSeek-V3 cohort's pre/post-emergence jump ($+0.11$) is far below the controls' same-split jump ($+0.35$), a matched difference-in-differences of $-0.24$; and the GPT-5 system-card cohort does not rise at all across its emergence ($-0.01$ raw, DiD $-0.17$), its hallucination-averse post-cutoff models correctly refusing on a name appearing only in a large contributor list.

Two implications follow. \emph{Recognition is not ingestion}: presence in the training corpus is necessary but far from sufficient, so crossing a cutoff does not manufacture recognition for an entity below the corpus-density threshold. And \emph{NameRank is not comparable across model vintages}: the vintage drift means longitudinal claims---including the predictions in Section~\ref{sec:predictions}---must be made on within-run \emph{relative} gaps, never absolute levels. The thresholded behavior has a mechanistic basis in how transformers store and retrieve facts: feed-forward layers act as key--value memories for factual associations~\citep{geva2021transformer}, with identifiable knowledge neurons per fact~\citep{dai2022knowledge, meng2022locating}; pretrained models function as implicit knowledge bases~\citep{petroni2019language} whose factual capacity scales with model size~\citep{roberts2020much}; accuracy on rare facts scales log-linearly with fact popularity~\citep{kandpal2023, mallen2023trust}; and popular knowledge actively suppresses less popular knowledge~\citep{overshadow2025}. Below the corpus-density threshold, silence rather than noise is what this retrieval stack predicts---which is what we observe.

\begin{table}[!htbp]
\centering
\small
\caption{Mean recognition for models whose training cutoff predates versus postdates each cohort's emergence date, with a matched difference-in-differences against stable controls. Both matched DiDs are $\leq 0$.}
\label{tab:cutoff-natexp}
\begin{tabular}{lrrrrr}
\toprule
Cohort & pre (refusal) & post (refusal) & raw jump & control & DiD \\
\midrule
DeepSeek-V3 authors & 0.01 & 0.12 & $+0.11$ & $+0.35$ & $\mathbf{-0.24}$ \\
GPT-5 system-card authors & 0.05 & 0.04 & $-0.01$ & $+0.16$ & $\mathbf{-0.17}$ \\
\bottomrule
\end{tabular}
\end{table}

\begin{table}[!htbp]
\centering
\small
\caption{Selected per-cohort slopes of per-model mean NameRank on training cutoff (per year), with Pearson $r$ over the $36$ models. Every established cohort rises at ${\sim}0.1$--$0.3$/yr; the recency cohorts are not steeper (drift removed in Table~\ref{tab:cutoff-natexp}).}
\label{tab:cutoff-slopes}
\begin{tabular}{lrr}
\toprule
Cohort & slope (/yr) & Pearson $r$ \\
\midrule
Foundation models & $+0.30$ & $+0.81$ \\
CS faculty (control) & $+0.30$ & $+0.67$ \\
OpenAlex researchers (control) & $+0.24$ & $+0.70$ \\
Reference set (control; incl.\ Hinton) & $+0.23$ & $+0.77$ \\
IMO gold & $+0.07$ & $+0.36$ \\
\textbf{DeepSeek-V3 authors} (recency) & $+0.04$ & $+0.18$ \\
\textbf{GPT-5 system-card authors} (recency) & $+0.02$ & $+0.20$ \\
\bottomrule
\end{tabular}
\end{table}

\section{Cross-Judge Robustness}
\label{app:cross-judge}

NameRank uses a single LLM judge (Section~\ref{sec:method-judge}). To test whether the findings depend on that choice, we re-judged a stratified sample of the panel's responses---spanning the diagnostic reference set, OpenAlex researchers, CS faculty, IMO gold medalists, GPT-5 system-card authors, writers, OSS projects, named methods, and foundation models---with two additional judges from different families (Claude Opus 4.6 and GPT-5.1), holding the entity contexts, gold answers, and the recognition rubric fixed. Because the verdict is binary, the natural agreement statistic is not a graded correlation but the rate at which two judges return the \emph{same} recognized/not-recognized decision on the same response, and Cohen's $\kappa$, which corrects that rate for chance agreement.

Two facts emerge, and together they say the findings do not depend on the judge. First, the three judges preserve the cohort ordering exactly: the silent$\to$discriminative$\to$universal ladder is monotone under each judge separately (Table~\ref{tab:cross-judge-bias}), so every relative comparison that constitutes a finding---the credential gap, the artifact-over-creator inversion, the institution gradient---is judge-invariant. Second, the judges differ in overall \emph{strictness} but not in ranking. Gemini and Claude are near-interchangeable (pairwise agreement $0.94$, Cohen's $\kappa=0.89$, and near-identical per-cohort levels), while GPT-5.1 is systematically more conservative: it credits recognition on $0.42$ of the sample against Gemini's $0.57$ and Claude's $0.58$, a roughly $0.15$ downward shift applied fairly uniformly across cohorts (Table~\ref{tab:cross-judge-corr}). That strictness lowers GPT-5.1's pairwise agreement with the other two ($\kappa \approx 0.67$--$0.69$) without reordering the cohorts---a stricter judge lowers all levels together---and because every headline result is a within-run relative gap read against the synthetic-null floor, a uniform level shift cancels.

\begin{table}[!htbp]
\centering
\small
\caption{Pairwise inter-judge agreement on the binary verdict over the $538$-response sample: fraction of matching decisions and Cohen's $\kappa$. Gemini and Claude are near-interchangeable; GPT-5.1 agrees on the ranking but is stricter.}
\label{tab:cross-judge-corr}
\begin{tabular}{lrr}
\toprule
Judge pair & \% agree & Cohen's $\kappa$ \\
\midrule
Gemini--Claude & 0.94 & 0.89 \\
Gemini--GPT-5.1 & 0.84 & 0.69 \\
Claude--GPT-5.1 & 0.83 & 0.67 \\
\midrule
Three-way unanimous & 0.81 & --- \\
\bottomrule
\end{tabular}
\end{table}

The ranking-preserved, level-shifted pattern is exactly what a robust instrument should show under a judge swap: the quantity that varies (absolute strictness) is the one every finding is designed to be invariant to, and the quantity that matters (the cohort ordering) is reproduced. Table~\ref{tab:cross-judge-bias} makes this concrete---the ladder is monotone down all three columns, and the near-uniform silent and universal cohorts (GPT-5 authors, named methods) sit at the extremes under every judge.

\begin{table}[!htbp]
\centering
\small
\caption{Per-cohort recognition rate under each judge on the shared sample. The silent$\to$universal ladder is monotone under all three judges; GPT-5.1 sits uniformly below Gemini and Claude, but the ordering is identical.}
\label{tab:cross-judge-bias}
\begin{tabular}{lrrr}
\toprule
Cohort & Gemini & Claude & GPT-5.1 \\
\midrule
GPT-5 system-card authors & 0.07 & 0.07 & 0.00 \\
IMO gold & 0.12 & 0.15 & 0.03 \\
OpenAlex long-tail researchers & 0.42 & 0.40 & 0.18 \\
CS faculty & 0.58 & 0.62 & 0.28 \\
Foundation models & 0.68 & 0.75 & 0.55 \\
OSS projects & 0.88 & 0.87 & 0.73 \\
Named methods & 0.95 & 0.93 & 0.77 \\
\midrule
\textbf{Overall} & \textbf{0.57} & \textbf{0.58} & \textbf{0.42} \\
\bottomrule
\end{tabular}
\end{table}

The headline findings are therefore not a single-judge artifact: the cohort ladder, the credential gap, and the silent zone all survive swapping the judge for a different model family. For a single-judge run Gemini and Claude are interchangeable; where absolute calibration matters, a stricter judge such as GPT-5.1 lowers all levels together without changing the ranking.

\section{Synthetic-Null Floor}
\label{app:synthetic}

To establish what NameRank assigns to a name absent from every pretraining corpus---the guessing floor---we construct fictional entities for each cohort recipe: plausible but non-existent researchers, founders, medalists, and papers, with verified-ungoogleable names, cohort-matched contexts, and gold answers built on the same recipe as the real cohort. Because the recognition judge credits only positively-verified facts, a model has nothing true to say about these entities, and the floor sits near zero (Table~\ref{tab:synthetic}): ${\leq}0.02$ for every people and artifact recipe. This is the property that makes a nonzero score meaningful---it is real recognition, not a guessing artifact.

One recipe floors higher and is handled explicitly. Fictional \emph{papers} floor at $0.06$, because a descriptive paper title (``\emph{Curriculum Distillation Under Label Scarcity\ldots}'') admits plausible guesses about the paper's topic and venue that the judge cannot always refute; this floor is subtracted where paper cohorts are compared. Fictional \emph{competition medalists}, by contrast, floor at essentially zero ($0.00$--$0.01$): the anti-confabulation provision (Section~\ref{sec:method-judge}) refuses to credit fine-grained memory of a specific medalist even against a real, well-documented event, so the floor-adjustment leaves every real credential cohort below the working-researcher baseline.

\begin{table}[!htbp]
\centering
\small
\caption{Synthetic-null recognition by recipe. Fictional people and artifacts earn essentially zero. Descriptive paper titles admit topic guesses (floor $\sim 0.06$); fictional competition medalists floor at essentially zero.}
\label{tab:synthetic}
\begin{tabular}{lrr}
\toprule
Synthetic recipe & $n$ & Recognition floor \\
\midrule
Founders, OSS, mid-tier, fellows & 16 & 0.01 \\
CS faculty & 10 & 0.01 \\
OpenAlex researchers & 6 & 0.02 \\
NOI medalists & 3 & 0.01 \\
IMO medalists & 6 & 0.00 \\
Papers (descriptive titles) & 4 & 0.06 \\
\midrule
People and artifacts (pooled) & 41 & \textbf{0.01} \\
\bottomrule
\end{tabular}
\end{table}

The residual floor is concentrated in the smaller open-weights models, which are more prone to fluent hallucination on unfamiliar names; the frontier and hallucination-averse models refuse or withhold recognition on the synthetics almost entirely. Recommended use: read every score against the $\sim 0.01$ people/artifact floor, and against $\sim 0.06$ for paper cohorts; differences below the relevant floor are not evidence of recognition.

\section{Confound Checks: Gold Length, Context, Wikipedia, and Attention}
\label{app:confounds}

Alternative explanations for the headline findings were tested and rejected: that the disambiguating context leaks the answer, that gold construction drives the credential gap, that NameRank reduces to Wikipedia presence, and that it reduces to a web-attention ranking. The first two are largely closed by the metric's design; the last two are empirical and rejected below.

\textbf{Context leakage and gold construction are closed by design.} The recognition verdict credits only facts that go \emph{beyond} the probe context and that are true of the specific entity (Section~\ref{sec:method-judge}). Two would-be confounds therefore cannot operate. A model cannot earn recognition by restating the disambiguating context---the fact that formerly inflated a graded coverage score now earns nothing---so context is an anchor, not a leak; the synthetic-null floor (Appendix~\ref{app:synthetic}), where fictional entities carry a full context yet score near zero, is the direct evidence. And because the verdict is binary on the presence of any one verified fact, gold-answer \emph{length} cannot scale it the way it scaled graded coverage: a longer gold offers more facts to hit but the entity is recognized on the first one---and empirically the synthetic-null floor is flat across gold lengths, so length buys no recognition without a real entity behind it.

\textbf{Wikipedia presence.} On the long-tail researcher cohort, a binary has-Wikipedia flag explains a modest fraction of recognition variance ($R^2 \approx 0.10$), well below the $h$-index's $R^2 \approx 0.22$, and it accounts for only a minority of the institution gradient. A binary flag also cannot rank the $92\%$ of researchers without an article, among whom NameRank still discriminates. NameRank tracks the accumulated corpus, not a Wikipedia-yes/no flag.

\begin{figure}[!htbp]
    \centering
    \includegraphics[width=\textwidth]{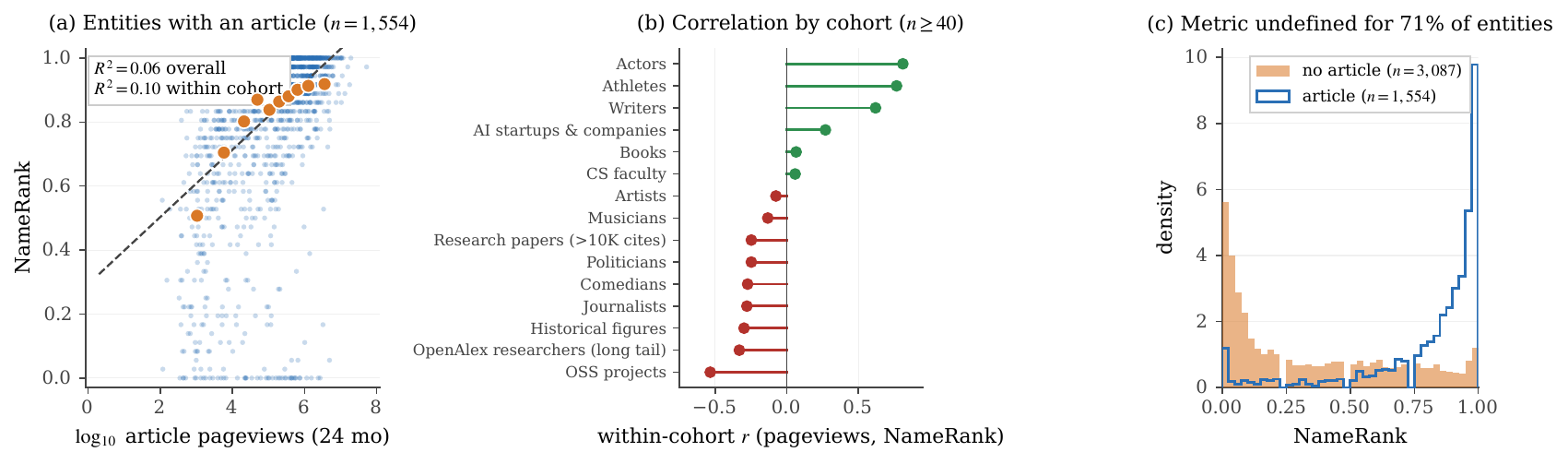}
    \caption{NameRank is not Wikipedia-pageview rank. \textbf{(a)}~Recognition against $\log_{10}$ article pageviews for entities with a resolvable article (small dots: entities; large markers: decile means). \textbf{(b)}~Within-cohort correlation: strong for celebrity cohorts (actors, athletes $r\approx0.85$), weak for technical and research cohorts ($r\approx0.05$--$0.48$). \textbf{(c)}~Recognition distribution for entities with and without an article; the metric is undefined for the $66\%$ without one.}
    \label{fig:attention}
\end{figure}

\textbf{Web attention.} Attention metrics also place entities on a single cross-domain axis, so we test the strongest freely available one: long-window en-Wikipedia article pageviews for every entity whose article a strict-disambiguation lookup resolves (Figure~\ref{fig:attention}). Two results reject the reduction. First, the metric is \emph{undefined} for the majority of entities---$66\%$ have no article at all---yet these span the entire discriminative zone where the credential, faculty, and long-tail findings live, so pageviews cannot even place most of the cohorts NameRank ranks. Second, where pageviews are defined they align with recognition only \emph{within the celebrity cohorts} (actors $r=0.83$, athletes $r=0.86$), where traffic and fame co-move by construction; inside the technical and research cohorts that carry the paper's findings the within-cohort correlation is weak (foundation models $r=0.48$, OSS $r=0.20$, named methods $r=0.23$, CS faculty $r=0.11$, long-tail researchers $r=0.05$). The strong cross-entity association ($R^2 \approx 0.20$ pooled) is thus carried by the celebrity block, not by the zone where NameRank does its work. Pageviews measure the \emph{flow} of current curiosity; NameRank measures the accumulated \emph{stock} of name-bearing text absorbed at training time (Section~\ref{sec:discussion-clock})---related axes, not the same one.

\section{Gender Gap by Cohort}
\label{app:gender}

First-name-inferred gender (via \texttt{gender-guesser}; ambiguous and unisex names dropped) for the person cohorts, restricted to cohorts with $\geq 5$ man- and woman-coded names. Table~\ref{tab:gender} reports the man-minus-woman NameRank gap.

\begin{table}[!htbp]
\centering
\small
\caption{Man-minus-woman NameRank gap by cohort. Among the research cohorts (top), the gap is a significant $+0.085$ for CS faculty but directional-only for the two smaller long-tail cohorts ($p>0.3$). The cross-profession cohorts (bottom) show larger raw gaps reflecting real-world fame, not a NameRank artifact---so the bias reading is restricted to the research comparison.}
\label{tab:gender}
\begin{tabular}{lrrrr}
\toprule
Cohort & $n_\text{man}$ & $n_\text{woman}$ & $\Delta$ (man$-$woman) & $p$ \\
\midrule
CS faculty & 252 & 77 & $+0.085$ & $0.004$ \\
OpenAlex long-tail researchers & 330 & 63 & $+0.025$ & $0.48$ \\
IKP long-tail researchers & 60 & 22 & $+0.055$ & $0.33$ \\
\midrule
Actors & 38 & 67 & $+0.21$ & $<10^{-3}$ \\
Athletes & 65 & 31 & $+0.27$ & $<10^{-3}$ \\
Writers & 62 & 80 & $+0.015$ & $0.38$ \\
Comedians & 20 & 16 & $+0.014$ & $0.86$ \\
\bottomrule
\end{tabular}
\end{table}

The research-cohort gap is significant for CS faculty ($+0.085$, $p=0.004$) and directional but not individually significant for the two smaller long-tail researcher cohorts ($+0.025$ and $+0.055$, $p>0.3$); the man-coded advantage is thus present in the best-powered cohort but modest and cohort-dependent, not a uniform law. The large cross-profession gaps (actor $+0.21$, athlete $+0.27$) are excluded from the bias reading because the sampled men and women in those cohorts differ in actual fame, and the near-zero writer and comedian gaps confirm the raw gaps track the sampling, not a NameRank artifact.

\textbf{Name-coding caveat.} \texttt{gender-guesser} infers gender from first names against a predominantly Western-name dictionary; it returns ``unknown'' or ``andy'' (androgynous) for most Chinese, Korean, and other romanized East-Asian given names, which are then dropped. The gendered sub-sample is therefore enriched for Western names, and the residual gender gap is partly entangled with the East-Asian-surname attenuation it sits beside---a name that codes as gender-unknown also tends to be one of the low-recognition non-Anglophone names. The $n_\text{woman}$ counts (e.g.\ $22$--$77$ across the researcher cohorts) reflect this attrition. We report the gap as an inherited corpus-and-judge property restricted to the controlled within-cohort comparison, not as a clean estimate of a pure gender effect net of name origin; disentangling the two would require a gender labeling that is accurate on CJK names (e.g.\ manual coding or a name-origin-aware classifier).